\documentclass[10pt,journal,compsoc]{IEEEtran}
\ifCLASSOPTIONcompsoc
  \usepackage[nocompress]{cite}
\else
  \usepackage{cite}
\fi
\ifCLASSINFOpdf
  \usepackage[pdftex]{graphicx}
\else
  \usepackage[dvips]{graphicx}
\fi
\usepackage{amsmath}
\usepackage{amssymb}

\usepackage[linesnumbered,ruled,vlined]{algorithm2e}

\usepackage{stfloats}
\newcommand\MYhyperrefoptions{bookmarks=true,bookmarksnumbered=true,
pdfpagemode={UseOutlines},plainpages=false,pdfpagelabels=true,
colorlinks=true,linkcolor={black},citecolor={black},urlcolor={black},
pdftitle={EPro-PnP: Generalized End-to-End Probabilistic Perspective-n-Points for Monocular Object Pose Estimation},%<!CHANGE!
pdfsubject={IEEE Transactions on Pattern Analysis and Machine Intelligence},%<!CHANGE!
pdfauthor={Hansheng Chen; Wei Tian; Pichao Wang; Fan Wang; Lu Xiong; Hao Li},%<!CHANGE!
pdfkeywords={Pose estimation; imaging geometry; probabilistic deep learning; 3D vision; Autonomous vehicles}}%<^!CHANGE!
\ifCLASSINFOpdf
\usepackage[\MYhyperrefoptions,pdftex]{hyperref}
\else
\usepackage[\MYhyperrefoptions,breaklinks=true,dvips]{hyperref}
\usepackage{breakurl}
\fi

\usepackage{booktabs}
\usepackage{listings}
\usepackage{multirow}
\usepackage{makecell}
\usepackage[dvipsnames]{xcolor}
\usepackage{mathtools}
\usepackage{etoolbox}

\SetKwInOut{KwIn}{Input}
\SetKwInOut{KwOut}{Output}
\SetKw{KwAnd}{and}

\SetCommentSty{mycommfont}
\SetAlgoNlRelativeSize{0}

\let\originalleft\left
\let\originalright\right
\renewcommand{\left}{\mathopen{}\mathclose\bgroup\originalleft}
\renewcommand{\right}{\aftergroup\egroup\originalright}
\DeclareMathOperator*{\argmax}{arg\,max}
\DeclareMathOperator*{\argmin}{arg\,min}
\newcommand{\diff}{\mathop{}\!\mathrm{d}}
\let\originalpartial\partial
\renewcommand{\partial}{\mathop{}\!\mathrm{\originalpartial}}
\newcommand{\expect}{\mathop{\mathbb{E}}}
\newcommand{\diag}{\mathop{\mathrm{diag}}}

\makeatletter
\patchcmd{\@algocf@start}% <cmd>
  {-1.5em}% <search>
  {-0.5em}% <replace>
  {}{}% <success><failure>
\def\@IEEEBIOskipN{2.4\baselineskip}
\makeatother

\begin{document}
\title{EPro-PnP: Generalized End-to-End Probabilistic\\ Perspective-n-Points for Monocular\\ Object Pose Estimation}

\author{Hansheng~Chen,
        Wei~Tian,
        Pichao~Wang,
        Fan~Wang,
        Lu~Xiong,
        and~Hao~Li% <-this % stops a space
\IEEEcompsocitemizethanks{
\IEEEcompsocthanksitem H. Chen is with the Department of Computer Science, Stanford University, Stanford, CA 94305 USA. Work done previously at Tongji University, Shanghai 201804, China. 
E-mail: hanshengchen@stanford.edu
\IEEEcompsocthanksitem W. Tian and L. Xiong are with the School of Automotive Studies, Tongji University, Shanghai 201804, China.
E-mail: \{tian\_wei,~xiong\_lu\}@tongji.edu.cn
\IEEEcompsocthanksitem P. Wang is with Amazon.com Inc, Seattle, WA 98109 USA. Work done previously at Alibaba Group (U.S.), Bellevue, WA 98004 USA.
E-mail: pichaowang@gmail.com
\IEEEcompsocthanksitem F. Wang is with Alibaba Group (U.S.), Sunnyvale, CA 94085 USA.
E-mail: fan.w@alibaba-inc.com
\IEEEcompsocthanksitem H. Li is with Artificial Intelligence Innovation and Incubation Institute, Fudan University, Shanghai 200433, China. Work done previously at Alibaba Group, Hangzhou 311121, China.
E-mail: lihao.lh@alibaba-inc.com}% <-this % stops a space
\thanks{(Corresponding author: Wei Tian.)}}


\markboth{Journal of \LaTeX\ Class Files,~Vol.~14, No.~8, August~2015}%
{Chen \MakeLowercase{\textit{et al.}}: EPro-PnP: Generalized End-to-End Probabilistic Perspective-n-Points for Monocular Object Pose Estimation}



\IEEEtitleabstractindextext{%
\begin{abstract}
Locating 3D objects from a single RGB image via Perspective-n-Point (PnP) is a long-standing problem in computer vision. 
Driven by end-to-end deep learning, recent studies suggest interpreting PnP as a differentiable layer, allowing for partial learning of 2D-3D point correspondences by backpropagating the gradients of pose loss. 
Yet, learning the entire correspondences from scratch is highly challenging, particularly for ambiguous pose solutions, where the globally optimal pose is theoretically non-differentiable w.r.t. the points.
In this paper, we propose the EPro-PnP, a probabilistic PnP layer for general end-to-end pose estimation, which outputs a distribution of pose with differentiable probability density on the SE(3) manifold. 
The 2D-3D coordinates and corresponding weights are treated as intermediate variables learned by minimizing the KL divergence between the predicted and target pose distribution. 
The underlying principle generalizes previous approaches, and resembles the attention mechanism.
EPro-PnP can enhance existing correspondence networks, closing the gap between PnP-based method and the task-specific leaders on the LineMOD 6DoF pose estimation benchmark. 
Furthermore, EPro-PnP helps to explore new possibilities of network design, as we demonstrate a novel deformable correspondence network with the state-of-the-art pose accuracy on the nuScenes 3D object detection benchmark. 
Our code is available at \url{https://github.com/tjiiv-cprg/EPro-PnP-v2}.
\end{abstract}

}

\maketitle

% is not selected <OR> if conference mode is selected - because compsoc
% conference papers position the abstract like regular (non-compsoc)
\IEEEdisplaynontitleabstractindextext

% \fi
\IEEEpeerreviewmaketitle

\ifCLASSOPTIONcompsoc
\IEEEraisesectionheading{\section{Introduction}\label{sec:introduction}}
\else
\section{Introduction}
\label{sec:introduction}
\fi

\IEEEPARstart{E}{stimating} the pose (i.e., position and orientation) of 3D objects from a single RGB image is an important problem in computer vision.
This field is often subdivided into specific tasks, e.g., 6DoF pose estimation for robot manipulation and 3D object detection for autonomous driving. Although they share the same fundamentals of pose estimation, the different nature of the data leads to biased choice of methods. Top performers~\cite{pgd, dd3d, detr3d} on the 3D object detection benchmarks~\cite{nuscenes,kitti} fall into the category of direct 4DoF pose prediction, leveraging the advances in end-to-end deep learning. On the other hand, the 6DoF pose estimation benchmark~\cite{linemod} is largely dominated by geometry-based methods~\cite{repose, DPOD}, which exploit the provided 3D object models and achieve a stable generalization performance. However, it is quite challenging to bring together the best of both worlds, i.e., training a geometric model to learn the object pose in an end-to-end manner.

\begin{figure}[!t]
\centering
\includegraphics[width=0.87\linewidth]{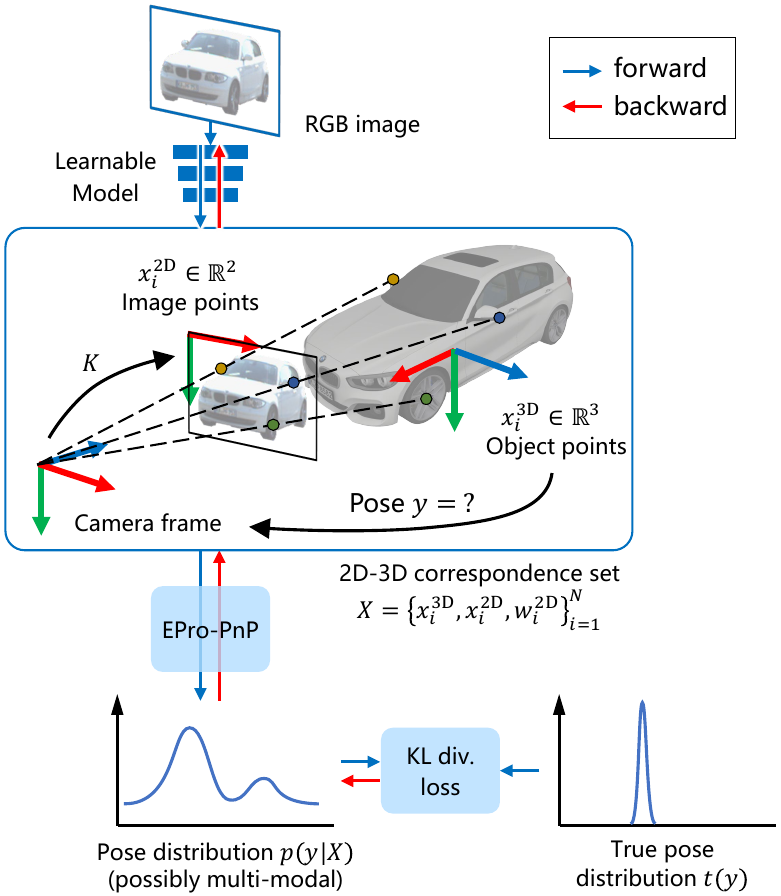}
\caption{An overview of the proposed framework. The predicted 2D-3D correspondences formulate a PnP problem. Instead of solving the optimal pose, EPro-PnP outputs a pose distribution, allowing the gradients of the KL loss w.r.t. the probability density to be backpropagated to train the correspondence network.}
\label{fig:header}
\end{figure}

There has been recent proposals for an end-to-end framework based on the Perspective-n-Point (PnP) approach~\cite{dsac, dsac++, blindpnp, BPnP}. The PnP algorithm itself solves the pose from a set of 3D points in object space and their corresponding 2D projections in image space, leaving the problem of constructing these correspondences.
Vanilla correspondence learning~\cite{pvnet, hybridpose, bb8, NOCS, pix2pose, DPOD, CDPN, RTM3D, deep-manta, pix2pose} leverages the geometric prior to build surrogate loss functions, forcing the network to learn a set of pre-defined correspondences. End-to-end correspondence learning~\cite{dsac, dsac++, blindpnp, BPnP} interprets the PnP solver as a differentiable layer and employs pose-driven loss function, so that gradient of the pose error can be backpropagated to the 2D-3D correspondences.

However, existing work on differentiable PnP learns only a portion of the correspondences (either 2D coordinates~\cite{BPnP}, 3D coordinates~\cite{dsac, dsac++} or corresponding weights~\cite{blindpnp}), assuming other components are given \emph{a priori}. This raises an important question: 
why not learn the entire set of points and weights altogether in an end-to-end manner?
Our intuition is: under such relaxed settings, the PnP problem could better describe pose ambiguity~\cite{manhardt2019, Schweighofer2006}, in the cases of symmetric objects~\cite{pix2pose} or uncertain observations. However, with the presence of ambiguity, the PnP problem has multiple local minima. Existing methods try to differentiate a point estimate of the pose (a single local minima), which is unstable in general, while the global optimum is neither easy to find nor differentiable.

To overcome the above limitations, we propose a generalized \textbf{e}nd-to-end \textbf{pro}babilistic \textbf{PnP} (EPro-PnP) module that enables learning the weighted 2D-3D point correspondences entirely from scratch.
The main idea is straightforward: a point estimate of pose is non-differentiable, but the probability density of pose is apparently differentiable, just like categorical classification scores. As shown in Figure~\ref{fig:header}, we interpret the output of PnP as a probabilistic distribution parameterized by the learnable 2D-3D correspondences. During training, the Kullback-Leibler (KL) divergence between the predicted and target pose distributions is minimized as the loss function, which can be efficiently calculated using the Adaptive Multiple Importance Sampling~\cite{amis} algorithm.

As a general approach, EPro-PnP inherently unifies existing correspondence learning techniques (Section~\ref{overview}). 
Moreover, just like the attention mechanism~\cite{vaswani2017attention}, the corresponding weights can be trained to automatically focus on important point pairs, allowing the networks to be designed with inspiration from attention-related work~\cite{detr,nonlocal,deformabledetr}.
% We therefore design the networks with inspiration from dense~\cite{detr, nonlocal} and deformable~\cite{deformabledetr} attention in object detection.
% , which fits dense correspondence~\cite{pix2pose, CDPN, DPOD, monorun} and a novel form of deformable correspondence, respectively.

To summarize, our main contributions are as follows: 
\begin{itemize}
    \item We propose the EPro-PnP, a probabilistic PnP layer for general end-to-end pose estimation with learnable 2D-3D correspondences, which can cope with pose ambiguity.
    \item We demonstrate that EPro-PnP can easily reach top-tier performance for 6DoF pose estimation by simply inserting it into the CDPN~\cite{CDPN} framework.
    \item We demonstrate the flexibility of EPro-PnP by proposing \emph{deformable correspondence learning} for accurate 3D object detection, where the entire 2D-3D correspondences are learned from scratch.
\end{itemize}

This extended paper presents new experiments with improved results and rigorous ablation studies. For 6DoF pose estimation on LineMOD, feeding 2D box size to the model has improved uncertainty handling, boosting pose accuracy to outperform RePOSE~\cite{repose}. New ablation studies reveal each loss's contribution and show that EPro-PnP can achieve competitive performance even without 3D models (B2 in Table~\ref{tab:cdpnablation}). For 3D object detection on nuScenes, EPro-PnP with an enhanced network now leads the field of single-frame image-based detectors, and the ablation studies highlight the importance of the Monte Carlo pose loss in handling ambiguous poses. Furthermore, we have also expanded our discussion on the derivative regularization loss.

\section{Related Work}

\subsection{Geometry-Based Object Pose Estimation}

In general, geometry-based methods exploit the points, edges or other types of representation that are subject to the projection constraints under the perspective camera. Then, the pose can be solved by optimization. A large body of work utilizes point representation, which can be categorized into sparse keypoints and dense correspondences. BB8~\cite{bb8} and RTM3D~\cite{RTM3D} locate the corners of the 3D bounding box as keypoints, while PVNet~\cite{pvnet} defines the keypoints by farthest point sampling and Deep MANTA~\cite{deep-manta} by handcrafted templates. On the other hand, dense correspondence methods \cite{pix2pose, CDPN, DPOD, monorun, NOCS} predict pixel-wise 3D coordinates within a cropped 2D region.
Most existing geometry-based methods follow a two-stage strategy, where the intermediate representations (i.e., 2D-3D correspondences) are learned with a surrogate loss function, which is sub-optimal compared to end-to-end learning.

\subsection{End-to-End Correspondence Learning}

To mitigate the limitation of surrogate correspondence learning, end-to-end approaches have been proposed to backpropagate the gradient from pose to intermediate representation. Using implicit differentiation w.r.t. the optimal pose or its approximations, Brachmann and Rother~\cite{dsac++} propose a dense correspondence network where 3D points are learnable, BPnP~\cite{BPnP} predicts 2D keypoint locations, and BlindPnP~\cite{blindpnp} learns the corresponding weight matrix given a set of unordered 2D/3D points. The above methods are all coupled with surrogate regularization loss, otherwise convergence is not guaranteed due to numerical instability~\cite{dsac++} and the non-differentiable nature of the optimal pose. Under the probabilistic framework, these methods can be regarded as a Laplace approximation approach (Section~\ref{overview}).

Beyond point correspondence, RePOSE~\cite{repose} proposes a feature-metric correspondence network trained by backpropagating the PnP solver (e.g. Levenberg-Marquardt), but it is insufficient under pose ambiguity although it can be leveraged as a local regularization technique in our framework (Section~\ref{localreg}).

\subsection{Probabilistic Deep Learning}

Probabilistic methods account for uncertainty in the model and the data, known respectively as epistemic and aleatoric uncertainty~\cite{kendall2017uncertainties}. The latter involves interpreting the prediction as learnable probabilistic distributions. Discrete categorical distribution via Softmax has been widely adopted as a smooth approximation of one-hot $\argmax$ for end-to-end classification. This inspired works such as DSAC~\cite{dsac}, a smooth RANSAC with a finite hypothesis pool. Meanwhile, tractable parametric distributions (e.g., normal distribution) are often used in predicting continuous variables~\cite{klloss, wu2020unsupervised, kendall2017uncertainties, VAE, Gilitschenski2020, monorun}, and mixture distributions can be employed to further capture ambiguity~\cite{makansi2019, Bishop94mixturedensity, Brachmann_2016_CVPR}, e.g., ambiguous 6DoF pose~\cite{bui20206d}. In this paper, we propose yet a unique contribution: backpropagating a complicated continuous distribution derived from a nested optimization layer (the PnP layer) approximated by importance sampling, essentially making it a continuous counterpart of Softmax.

\section{Generalized End-to-End Probabilistic PnP}

\subsection{Overview} \label{overview}

Given an object proposal, our goal is to predict a set $X = \left\{x^\text{3D}_i,x^\text{2D}_i,w^\text{2D}_i\right\}_{i=1}^N$ of $N$ corresponding points, with 3D object coordinates $x^\text{3D}_i \in \mathbb{R}^3$, 2D image coordinates $x^\text{2D}_i \in \mathbb{R}^2$, and 2D weights $w^\text{2D}_i \in \mathbb{R}^2_+ $, from which a weighted PnP problem can be formulated to estimate the object pose relative to the camera.

The essence of a PnP layer is searching for an optimal pose $y$ (expanded as rotation matrix $R$ and translation vector $t$) that minimizes the cumulative squared weighted reprojection error:
\begin{equation}
\smash[b]{\argmin_{y} \frac{1}{2} \sum_{i=1}^N \left\| \smash[b]{\underbrace{w_i^\text{2D} \circ \left( \pi(Rx_i^\text{3D} + t) - x_i^\text{2D} \right)}_{f_i(y) \in \mathbb{R}^2}} \right\|^2,}
\vphantom{\underbrace{\left((O_o^O)\right)}_{o}}
\label{eqn:basicpnp}
\end{equation}
where $\pi(\cdot)$ is the projection function with camera intrinsics involved, $\circ$ stands for element-wise product, and $f_i(y)$ compactly denotes the weighted reprojection error.

Eq.~(\ref{eqn:basicpnp}) formulates a non-linear least squares problem that may have non-unique solutions, i.e., pose ambiguity~\cite{manhardt2019, Schweighofer2006}. Previous work~\cite{dsac++, BPnP, blindpnp} only backpropagates through a local solution $y^\ast$, which is inherently unstable and non-differentiable.
% where the $\argmin$ function is not continuous, let alone differentiable. The continuity breaks typically when the global optimum switches from one mode to another, due to pose ambiguity~\cite{manhardt2019, Schweighofer2006}. 
% Previous efforts~\cite{dsac++, BPnP, blindpnp} on backwarding through the PnP operation only derives a single local optimum, which does not guarantee global convergence.
To construct a differentiable alternative for end-to-end learning, we model the PnP output as a distribution of pose, which guarantees differentiable probability density. The cumulative error is considered to be the negative logarithm of the likelihood function $p(X|y)$ defined as:
\begin{equation}
p \left(X \middle| y \right) = \exp -\frac{1}{2} \sum_{i=1}^N \left\| f_i(y) \right\|^2 .
\label{nll}
\end{equation}
With an additional prior pose distribution $p(y)$, we can derive the posterior pose $p(y|X)$ via the Bayes theorem. Using an \emph{uniform prior} in the domain $Y$, the posterior density is simplified to the normalized likelihood:
\begin{equation}
p(y|X)
% = \frac{p(X|y)}{ \int p(X|y) \diff{y}}
= \frac{\exp -\frac{1}{2} \sum_{i=1}^N \left\| f_i(y) \right\|^2}{ \int_Y \exp -\frac{1}{2} \sum_{i=1}^N \left\| f_i(y) \right\|^2 \diff{y}}.
\label{posterior}
\end{equation}
Eq.~(\ref{posterior}) can be interpreted as a continuous counterpart of categorical Softmax.

\subsubsection{KL Loss Function}
During training, given a target pose distribution with probability density $t(y)$, the KL divergence $D_\text{KL}\left(t(y) \| p(y|X) \right)$ is minimized as training loss. Intuitively, pose ambiguity can be captured by the multiple modes of $p(y|X)$, and convergence is ensured such that wrong modes are suppressed by the loss function. Substituting Eq.~(\ref{posterior}), the KL divergence loss can be re-written as follows:
\begin{align}
\mathcal{L}_\text{KL} &= \int_Y t(y) \left(\log{t(y)} - \log{p(y|X)}\right) \diff{y} \notag\\
&= -\int_Y t(y) \log{\frac{p(X|y)}{\int_Y p(X|y)\diff{y}}} \diff{y} + \mathit{const} \notag\\
&= -\int_Y t(y) \log{p(X|y)} \diff{y} + \log{\int_Y p(X|y)\diff{y}} + \mathit{const}.
\end{align}
In practice, we drop the constant relevant to the target distribution so that it is effectively a cross-entropy loss.
In addition, we empirically find it effective to set a narrow (Dirac-like) target distribution centered at the ground truth $y_\text{gt}$, yielding the simplified loss (after substituting Eq.~(\ref{nll})):
\begin{equation} 
\mathcal{L}_\text{KL} = \underbrace{\frac{1}{2} \sum_{i=1}^N \left\| f_i(y_\text{gt}) \right\|^2}_{\mathclap{\mathcal{L}_\text{tgt} \text{ (reproj. at target pose)}}} + \underbrace{\log \int_Y \exp - \frac{1}{2} \sum_{i=1}^N \left\| f_i(y) \right\|^2 \diff{y}}_{\mathcal{L}_\text{pred}\text{ (reproj. at predicted pose)}}. 
\label{symbloss}
\end{equation}
The only remaining problem is the integration in the second term, which is elaborated in Section~\ref{mcloss}.

\begin{figure}[t]
\begin{center}
    \includegraphics[width=0.9\linewidth]{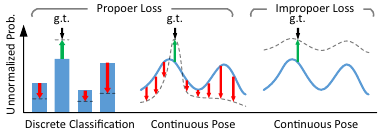}
\end{center}
\vspace{-1ex}
\caption{Learning a discrete classifier vs. Learning the continuous pose distribution. A discriminative loss function (left) shall encourage the unnormalized probability for the correct prediction as well as penalize for the incorrect. 
A one-sided loss (right) will degrade the distribution if the model is not well-regularized.}
\label{fig:distr}
\end{figure}

\subsubsection{Comparison to Reprojection-Based Method}
\label{reproj}
The two terms in Eq.~(\ref{symbloss}) are concerned with the reprojection errors at target and predicted pose respectively. The former is often used as a surrogate loss in previous work~\cite{BPnP, dsac++, monorun}. However, the first term alone cannot handle learning all 2D-3D points without imposing strict regularization, as the minimization could simply collapse all the 2D-3D points. The second term originates from the normalization factor in Eq.~(\ref{posterior}), and is crucial to a discriminative loss function, as shown in Figure~\ref{fig:distr}.

\subsubsection{Comparison to Implicit Differentiation Method} 
Existing work on end-to-end PnP~\cite{BPnP, blindpnp}
% backpropagates the gradient \wrt $y^\ast$ by
derives a single solution of a particular solver $y^\ast = \mathit{PnP}(X)$ via implicit function theorem~\cite{declarative}, assuming $\nabla_y \frac{1}{2} \sum_{i=1}^N \left\| f_i(y) \right\|^2 \negmedspace \bigm|_{y=y^\ast} = 0$. In the probabilistic framework, this is essentially the Laplace method that approximates the posterior by $\mathcal{N}(y^\ast, \Sigma_{y^\ast})$, where both $y^\ast$ and $\Sigma_{y^\ast}$ can be estimated by the PnP solver with analytical derivatives~\cite{monorun}. 
% Therefore, the KL divergence loss is analytically tractable.
If $\Sigma_{y^\ast}$ is simplified to be isotropic, the approximated KL divergence can be simplified into the L2 loss $\|y^\ast - y_\text{gt}\|^2$ used in \cite{blindpnp}. However, the Laplace approximation is inaccurate for non-normal posteriors with ambiguity, therefore does not guarantee global convergence. Besides, implicit differentiation itself may be prone to numerical instability~\cite{dsac++}.

\subsection{Monte Carlo Pose Loss} \label{mcloss}

In this section, we introduce a GPU-friendly efficient Monte Carlo approach to the integration in the proposed loss function, based on the Adaptive Multiple Importance Sampling (AMIS) algorithm~\cite{amis}.
% , which can be easily implemented for efficient parallel computation on GPU.

Considering $q(y)$ to be the probability density function of a proposal distribution that approximates the shape of the integrand $\exp -\frac{1}{2} \sum_{i=1}^N \left\| f_i(y) \right\|^2$, and $y_j$ to be one of the $K$ samples drawn from $q(y)$, the estimation of the second term $\mathcal{L}_\text{pred}$ in Eq.~(\ref{symbloss}) is thus:
\begin{equation}
 \mathcal{L}_\text{pred} \approx \log \frac{1}{K} \sum_{j=1}^K \underbrace{\frac{\exp - \frac{1}{2} \sum_{i=1}^N \left\| f_i(y_j)\right\|^2}{q(y_j)}}_{v_j \text{ (importance weight)}},
\label{vanillais}
\end{equation}
where $v_j$ compactly denotes the importance weight at $y_j$. Eq.~(\ref{vanillais}) gives the vanilla importance sampling, where the choice of proposal $q(y)$ strongly affects the numerical stability. The AMIS algorithm is a better alternative as it iteratively adapts the proposal to the integrand. 

In brief, AMIS utilizes the sampled importance weights from past iterations to estimate the new proposal. Then, all previous samples are re-weighted as being homogeneously sampled from a mixture of the overall sum of proposals.~\cite{amis} Initial proposal can be determined by the mode and covariance of the predicted pose distribution (see supplementary for details). A pseudo-code is given below.

\setlength{\interspacetitleruled}{0pt}%
\setlength{\algotitleheightrule}{0pt}%
\begin{algorithm}[h]
\DontPrintSemicolon
  \KwIn{$X = \{x_i^\text{3D}, x_i^\text{2D}, w_i^\text{2D}\}_{i=1}^N$}
  \KwOut{$\mathcal{L}_\text{pred}$}
  $y^\ast, \Sigma_{y^\ast} \gets \mathit{PnP}(X)$ \tcp*{Laplace approximation}
  Fit $q_1(y)$ to $y^\ast, \Sigma_{y^\ast}$ \tcp*{initial proposal}
  
  \For{$1 \le t \le T$}{
    Generate $K^\prime$ samples $y^t_{j=1 \cdots K^\prime}$ from $q_t(y)$
    
    \For{$1 \le j \le K^\prime$}{
      $P_j^t \gets p(X|y_j^t)$ \tcp*{evaluate integrand}
    }
    
    \For{$1 \le \tau \le t$ \KwAnd $1 \le j \le K^\prime$}{
      $Q_j^\tau \gets \frac{1}{t}\sum_{m=1}^t q_m(y_j^\tau)$ \tcp*{evaluate proposal mix}
     
      $v_j^\tau \gets P_j^\tau / Q_j^\tau$  \tcp*{importance weight}
    }
    
    \If{$t < T$}{
    Estimate $q_{t+1}(y)$ from all weighted samples $\{y_j^\tau, v_j^\tau\ | \, 1 \le \tau \le t, 1 \le j \le K^\prime\}$}
  }
  $\mathcal{L}_\text{pred} \gets \log \frac{1}{T K^\prime} \sum_{t=1}^{T} \sum_{j=1}^{K^\prime} v_j^t$
\end{algorithm}

In this paper, we empirically set the AMIS iteration count $T$ to 4, and the number of samples per iteration $K^\prime$ to 128 for 6DoF pose and 32 for 4DoF pose (1D yaw-only orientation). These hyperparameters can be adjusted to balance computation and accuracy.

\subsubsection{Choice of Proposal Distribution} 

We use separate proposal distributions for position and orientation, as the orientation space is non-Euclidean. For position, we adopt the 3DoF multivariate t-distribution.
% as in the original AMIS paper.
For 1D yaw-only orientation, we use a mixture of von Mises and uniform distribution. For 3D orientation represented by unit quaternion, the angular central Gaussian distribution~\cite{ACG} is adopted. 
% Details on the parameter estimation are given in the supplementary materials.

\subsection{Backpropagation} \label{backprop}

Although backpropagation can be simply implemented with automatic differentiation packages, here we analyze the gradients of the loss function for an intuitive understanding of the learning process. In general, the gradients of the loss function defined in Eq.~(\ref{symbloss}) is:
\begin{equation}
\nabla\mathcal{L}_\text{KL} = \nabla \frac{1}{2} \sum_{i=1}^N \left\| f_i(y_\text{gt}) \right\|^2 -
\hspace{-1ex} \expect_{y \sim p(y|X)}{\hspace{-1ex} \nabla \frac{1}{2} \sum_{i=1}^N \left\| f_i(y) \right\|^2},
\label{derivative}
\end{equation}
where the first term is the gradient of reprojection errors at target pose, and the second term is the expected gradient of reprojection errors over predicted pose distribution, which is approximated by backpropagating the importance weights in the Monte Carlo pose loss.

\subsubsection{Balancing Uncertainty and Discrimination} 

\begin{figure}[t]
\begin{center}
    \includegraphics[width=0.98\linewidth]{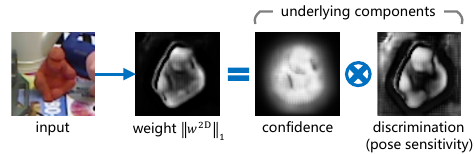}
\end{center}
\vspace{-1ex}
\caption{The learned corresponding weight can be factorized into inverse uncertainty and discrimination. 
Typically, inverse uncertainty roughly resembles the foreground mask, while discrimination emphasizes the 3D extremities of the object.
% , just as a 2D detector locates the object by its 2D extremities.
}
\label{fig:balance}
\end{figure}

Consider the negative gradient w.r.t. the corresponding weights $w_i^\text{2D}$:
\begin{equation}
-\nabla_{w_i^\text{2D}} \mathcal{L}_\text{KL} = w_i^\text{2D} \circ \left( -r_i^{\circ 2}(y_\text{gt}) + \hspace{-1ex} \expect_{y \sim p(y|X)}{\hspace{-1ex}r_i^{\circ 2}(y)} \right),
\end{equation}
where $r_i(y) = \pi(Rx_i^\text{3D} + t) - x_i^\text{2D}$ (unweighted reprojection error), and $(\cdot)^{\circ 2}$ stands for element-wise square. The first bracketed term $-r_i^{\circ 2}(y_\text{gt})$ with negative sign indicates that correspondences with large reprojection error (hence high uncertainty) shall be weighted less. The second term $\expect_{y \sim p(y|X)}{r_i^{\circ 2}(y)}$ is relevant to the variance of reprojection error over the predicted pose. The positive sign indicates that correspondences sensitive to pose variation should be weighted more, because they provide stronger pose discrimination. The final gradient is thus a balance between the uncertainty and discrimination, as shown in Figure~\ref{fig:balance}. Existing work~\cite{monorun, pvnet} on learning uncertainty-aware correspondences only considers the former, hence lacking the discriminative ability.

\subsection{Limitations and Derivative Regularization Loss} \label{localreg}

In practice, we observe that the KL divergence loss has two limitations:
\begin{itemize}
    \item While the KL divergence is a good metric for the probabilistic distribution, existing evaluation protocols are all based on the point estimate of pose $y^\ast$. Therefore, for inference it is still required to locate a mode $y^\ast$ of the posterior $p(y|X)$ by solving the PnP problem in Eq.~(\ref{eqn:basicpnp}), which could be sub-optimal if trained solely with the KL loss.
    \item The 2D-3D correspondences are underdetermined if we only impose the KL loss when training the network. Learning these entangled elements could be difficult if the network architecture is not designed carefully with preferable inductive bias.
\end{itemize}

The above limitations can be mitigated by an additional regularization loss on $y^\ast$ that backpropagates through the Gauss-Newton (GN) least squares solver or its variants~\cite{repose}. We call it the \emph{derivative regularization loss}, since GN is a derivative-based optimizer, and the loss therefore acts on the derivatives of the log-density $\log{p(y|X)}$ to direct the GN increment $\Delta y$ towards the true pose $y_\text{gt}$.

To employ the regularization during training, a detached solution $y^\ast$ is obtained first. Then, at $y^\ast$, a final GN increment is evaluated (which ideally equals 0 if $y^\ast$ has already converged to the local optimum):
\begin{equation}
    \Delta y = -{\underbrace{(J^\text{T}J)}_{\mathclap{H\text{ (approx.)}}}}^{-1}\underbrace{J^\text{T} F(y^\ast)}_{g},
    \label{gnstep}
\end{equation}
where $F(y^\ast) = \left[f_1^\text{T}(y^\ast), f_2^\text{T}(y^\ast), \cdots, f_N^\text{T}(y^\ast)\right]^\text{T}$ is the flattened weighted reprojection errors of all points, $J = \partial{F(y)} / \partial{y^\text{T}} \negmedspace \bigm|_{y=y^\ast}$ is the Jacobian matrix, $J^\text{T} F(y)$ equals the gradient $g$ of the negative log-likelihood (NLL) w.r.t. object pose, i.e., $\partial{\frac{1}{2} \sum_{i=1}^N \left\| f_i(y) \right\|^2} / \partial{y}$, and $J^\text{T}J$ is an approximation of the Hessian matrix $H=\partial{g} / \partial{y^\text{T}}$. We therefore design the regularization loss as follows:
\begin{equation}
\mathcal{L}_\text{reg} = l(y^\ast + \Delta y, y_\text{gt}),
\label{regloss}
\end{equation}
where $l(\cdot, \cdot)$ is a distance metric for pose. We adopt smooth L1 for position and cosine similarity for orientation (see supplementary materials for details). Note that the gradient is only backpropagated through $\Delta y$, which is analytically differentiable w.r.t. the 2D-3D correspondences. 

This loss not only addresses the first limitation by moving $y^\ast$ towards $y_\text{gt}$, but also partially disentangles the 2D-3D correspondences. To analyze the effect of the loss on the correspondences, we consider a local approximation of Eq.~(\ref{regloss}), assuming equal weights for position and orientation:
\begin{align}
\mathcal{L}_\text{reg} &\approx \left\| y^\ast + \Delta y - y_\text{gt} \right\|^2 \notag\\
&= \left\| y^\ast -\smash[b]{\underbrace{(J^\text{T}J)^{-1} J^\text{T}}_{J^+}} F(y^\ast) - y_\text{gt} \right\|^2. \vphantom{\underbrace{(J^\text{T}J)^{-1} J^\text{T}}_{J^+}}
\end{align}
Note that $(J^\text{T}J)^{-1} J^\text{T}$ is also the pseudo inverse of the matrix $J$, which can be denoted by $J^+$ for brevity. Then, taking the first-order approximation  $F(y^\ast) = F(y_\text{gt}) + J\left( y^\ast - y_\text{gt} \right)$, the loss can be approximated into:
\begin{align}
\mathcal{L}_\text{reg} &\approx \left\| y^\ast - J^+ \left(F(y_\text{gt}) + J\left( y^\ast - y_\text{gt} \right)\right) - y_\text{gt} \right\|^2 \notag\\
&= \left\| J^+ F(y_\text{gt}) \right\|^2.
% \notag\\
% &= F^\text{T}(y_\text{gt}) \left( JJ^\text{T} \right)^+ F^\text{T}(y_\text{gt})
\end{align}
This indicates that the derivative regularization loss is analogous to the reprojection-based surrogate loss $\left\|F(y_\text{gt}) \right\|^2$ (Section~\ref{reproj}). Although the extra weighting matrix $J^+$ makes the individual elements in the reprojection vector $F(y_\text{gt})$ underdetermined, over multiple samples and mini-batches there remains a tendency of independently minimizing each of the elements, i.e., minimizing the reprojection error of each correspondence. Thus, it helps to overcome the potential training difficulties associated with the KL loss.

The regularization loss can also serve as an independent objective for training pose estimators, akin to RePOSE~\cite{repose}. However, since we observe that this objective alone is not effective in addressing pose ambiguity, it is treated as a secondary regularization in this study.

\section{Implementation Details}

\subsection{Dynamic KL Loss Weight}

Following \cite{monorun}, we compute a dynamic loss weight for $\mathcal{L}_\text{KL}$ so that the magnitude of its gradients is consistent regardless of the entropy of the distribution. This is implemented by computing the exponential moving average (EMA) of the 1-norm of the sum of weights $\left\| \sum_{i=1}^{N}{w^\text{2D}_i} \right\|_1$, and using the reciprocal of the EMA value as the dynamic loss weight for $\mathcal{L}_\text{KL}$. Intuitively, this cancels out the effect of the magnitude of $w^\text{2D}_i$ on the loss gradients w.r.t. $x^\text{2D}_i$ and $x^\text{3D}_i$.

\subsection{Adaptive Huber Kernel}

For the PnP formulation in Eq.~(\ref{eqn:basicpnp}), the plain L2 reprojection errors $\left\| f_i(y) \right\|^2$ are sensitive to outliers, which limits the model's expressiveness in representing multi-modal distributions that characterizes ambiguity. Therefore, we robustify the reprojection errors using the Huber kernel $\rho(\cdot)$, yielding an alternative formulation:
\begin{equation}
\argmin_{y} \frac{1}{2} \sum_{i=1}^N \rho \left( \left\| f_i(y) \right\|^2 \right).
\label{eqn:robustpnp}
\end{equation}
The Huber kernel with threshold $\delta$ is defined as:
\begin{equation}
 \rho(s) = 
 \begin{dcases}
 s, & s \leq \delta^2,\\
 \delta(2 \sqrt{s} - \delta), & s > \delta^2.
 \end{dcases}
\label{eqn:huber}
\end{equation}
To robustify the weighted reprojection errors of various scales, we adopt an adaptive threshold $\delta$ defined as a function of the weights ${w^\text{2D}_i}$ and 2D coordinates ${x^\text{2D}_i}$:
\begin{equation}
    \delta = \delta_\text{rel} \frac{\left\| \bar{w}^\text{2D} \right\|_1}{2}  \left( \frac{1}{N-1}\sum_{i=1}^N{\left\|x^\text{2D}_i - \bar{x}^\text{2D}\right\|^2} \right)^{\negthickspace\frac{1}{2}}
\end{equation}
with the relative threshold $\delta_\text{rel}$ as a hyperparameter, and the mean vectors $\bar{w}^\text{2D}=\frac{1}{N}\sum_{i=1}^N{w_i^\text{2D}},\, \bar{x}^\text{2D}=\frac{1}{N}\sum_{i=1}^N{x_i^\text{2D}}$.

Accordingly, the reprojection errors $F(y)$ and Jacobian matrix $J$ in Eq.~(\ref{gnstep}) have to be rescaled (see supplementary).

\subsection{Initialization}
\label{init}

Since the LM solver only finds a local solution, initialization plays a determinant role in dealing with ambiguity. We implement a random sampling algorithm analogous to RANSAC, to search for the global optimum efficiently.

Given the $N$-point correspondence set $X = \left\{x^\text{3D}_i,x^\text{2D}_i,w^\text{2D}_i\right\}_{i=1}^N$, we generate $M$ subsets consisting of $n$ corresponding points each ($3 \leq n < N$), by repeatedly sub-sampling $n$ indices without replacement from a multinomial distribution, whose probability mass function $p(i)$ is defined by the corresponding weights:
\begin{equation}
    p(i) = \frac{\left\|w^\text{2D}_i\right\|_1}{\sum_{i=1}^N{\left\|w^\text{2D}_i\right\|_1}}.
\end{equation}
From each subset, a pose hypothesis can be solved via the LM algorithm within very few iterations (e.g. 3 iterations). This is implemented as a batch operation on GPU, and is rather efficient for small subsets. We take the hypothesis with maximum log-likelihood $\log{p(X|y)}$ as the initial point, starting from which subsequent LM iterations are computed on the full set $X$. 

\subsubsection{Training Mode Initialization}
During training, the LM PnP solver is utilized for estimating the location and concentration of the initial proposal distribution in the AMIS algorithm. The location is very important to the stability of Monte Carlo training. If the LM solver fails to find the global optimum, and the location of the local optimum is far from the true pose $y_\text{gt}$, the balance between the two opposite signed terms in Eq.~(\ref{symbloss}) may be broken, leading to exploding gradient in the worst case scenario. To avoid such problem, we adopt a simple initialization trick: we compare the log-likelihood $\log{p(X|y)}$ of the ground truth $y_\text{gt}$ and the selected hypothesis, and then keep the one with higher likelihood as the initial state of the LM solver. 

\section{6DoF Pose Estimation based on CDPN}

To demonstrate that EPro-PnP can be applied to off-the-shelf 2D-3D correspondence networks, experiments have been conducted on CDPN~\cite{CDPN}, a dense correspondence network for 6DoF pose estimation.

\subsection{Network Architecture}

The original CDPN feeds cropped image regions within the detected 2D boxes into the pose estimation network, to which two decoupled heads are appended for rotation and translation respectively.
The rotation head is PnP-based while the translation head uses explicit center and depth regression. 
This paper discards the translation head to focus entirely on PnP, and modifies only the last layer of the rotation head for strict comparison to the baseline.

\begin{figure}[!t]
\centering
\includegraphics[width=0.85\linewidth]{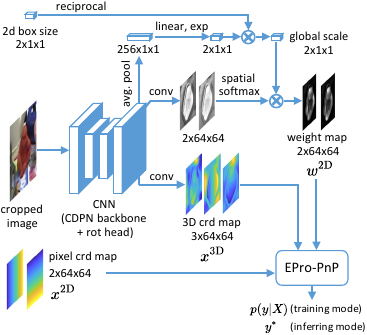}
% \vspace{-1ex}
\caption{The 6DoF pose estimation network modified from CDPN~\cite{CDPN}, with spatial Softmax and global weight scaling.}
\label{fig:cdpnnet}
\end{figure}

As shown in Fig.~\ref{fig:cdpnnet}, apart from the standard 3D coordinate map, the network predicts a 2-channel weight map (originally it is a single channel segmentation mask). We find it necessary to predict a global scale $w^\text{2D}_\text{S}$ separately, and apply it to the normalized weights ${w^\text{2D}_\text{N}}_{\hspace{-.3em} i}$ that satisfies $\sum_{i=1}^N{{w^\text{2D}_\text{N}}_{\hspace{-.3em} i}}=[1, 1]^\text{T}$. Intuitively, the global scale controls the entropy of the pose distribution $p(y|X)$ as it scales the entire log-likelihood, while the normalized weights determines the relative importance of each correspondence. This helps to overcome the entangling effect of the KL loss mentioned in Section~\ref{localreg}. Inspired by the attention mechanism~\cite{vaswani2017attention}, the normalized weights are activated via spatial Softmax, focusing on important regions in the image. The global scale is usually inversely proportional to the 2D size of the object due to the uncertainty in reprojection, and is hard-coded as such in this network.

The original CDPN imposes masked coordinate regression loss\cite{CDPN} to learn the dense correspondences, using the ground truth object 3D models to render the target masks and 3D coordinate maps. With EPro-PnP, however, this extra geometry supervision is optional, as we demonstrate that the entire network can be trained solely by the KL loss $\mathcal{L}_\text{KL}$ and/or the derivative regularization loss $\mathcal{L}_\text{reg}$. To reduce the Monte Carlo overhead, 512 points are randomly sampled from the 64\texttimes64 dense points to compute $\mathcal{L}_\text{KL}$.

\begin{table}[!t]
\caption{Results of the CDPN baseline. A0 and A1 are reproduced with the official code (\url{https://git.io/JXZv6}).}
\label{tab:cdpnbaseline}
\centering
    \scalebox{1.0}{%
    \setlength{\tabcolsep}{0.4em}
    \begin{tabular}{clcccl}
        \toprule
        \multirow{2}[2]{*}{ID} & \multirow{2}[2]{*}{Method} & \multicolumn{3}{c}{ADD(-S)} & \multirow{2}[2]{*}{Mean} \\
        \cmidrule(lr){3-5}
        {} & {} & 0.02d & 0.05d & 0.1d \\
        \midrule
        A0 & CDPN-Full~\cite{CDPN} & 29.10 & 69.50 & 91.03 & 63.21 \\
        A1 & CDPN w/o trans. head & 15.93 & 46.79 & 74.54 & 45.75 \\
        A2 & A1 \textrightarrow\ Batch=32, LM solver & 21.17 & 55.00 & 79.96 & 52.04 \\
        \bottomrule
    \end{tabular}}
\end{table}

\subsection{Dataset and Metrics}
As in CDPN, we use the LineMOD~\cite{linemod} 6DoF pose estimation dataset to conduct our experiments. The dataset consists of 13 sequences, each containing about 1.2K images annotated with 6DoF poses of a single object. Following \cite{Brachmann_2016_CVPR}, the images are split into the training and testing sets, with about 200 images per object for training. For data augmentation, we use the same synthetic data as in CDPN~\cite{CDPN}.
% , in which 1000 images per object are rendered with random background images from the PASCAL VOC2012 dataset.

We use two common metrics for evaluation: ADD(-S) and $n\text{\textdegree}, n\,\text{cm}$. The ADD measures whether the average deviation of the transformed model points is less than a certain fraction of the object’s diameter (e.g., ADD-0.1d). For symmetric objects, ADD-S computes the average distance to the closest model point. $n\text{\textdegree}, n\,\text{cm}$ measures the accuracy of pose based on angular/positional error thresholds. All metrics are presented as percentages. 

Despite that some objects in the dataset are nearly rotational symmetric, we observe that our model has no trouble identifying their exact orientations. Therefore, the presented results shall be closer to the scenario without pose ambiguity.

\subsection{Baseline}

For strict comparison, general settings are kept the same as in CDPN~\cite{CDPN} (with ResNet-34~\cite{resnet} as backbone). As shown in Table~\ref{tab:cdpnbaseline}, the original CDPN-Full (A0) trains the network in 3 stages totaling 480 epochs using RMSprop. With the translation head removed, we only train the rotation head in a single stage of 160 epochs (A1), which greatly impacts the pose accuracy (45.75 vs. 63.21). Additionally, we improve the baseline by using the LM solver with Huber kernel at test time, and increase the batch size to 32 for less training wall time (A2). Instead of using the advanced initialization technique in Section~\ref{init}, we adopt the simple EPnP~\cite{EPnP} initialization without RANSAC.

\subsection{Main Results and Discussions}
As shown in Table~\ref{tab:cdpnablation}, we conduct ablation studies to reveal the contributions of the Monte Carlo KL loss $\mathcal{L}_\text{KL}$, the derivative regularization loss $\mathcal{L}_\text{reg}$, the original coordinate regression loss $\mathcal{L}_\text{crd}$ in CDPN~\cite{CDPN}, and initializing the model with pretrained weights from A1.

\subsubsection{KL Loss vs. Coordinate Regression}
Training the model from scratch with the KL loss alone (B0) significantly outperforms the baseline model (A2) trained with the coordinate regression loss (61.87 vs. 52.04), despite the lack of geometry supervision from the ground truth object 3D models.

\subsubsection{KL Loss and Derivative Regularization}
Both the KL loss (B0) and the derivative regularization loss (B1) performs well independently on this benchmark. Because pose ambiguity is not noticeable in LineMOD dataset, the solver-based derivative regularization loss performs better than the KL loss (63.15 vs. 61.87). 
Nevertheless, the best possible pose accuracy without knowing the object geometry can be achieved when combining both loss functions together (B2), even outperforming CDPN-Full (A0) by a clear margin (67.36 vs. 63.21).

\subsubsection{With Knowledge of the Object 3D Models}

On top of B2, one can further impose the coordinate regression loss $\mathcal{L}_\text{crd}$ (B4) with target 3D coordinates rendered from the object 3D models, further improving the pose accuracy. Yet a better approach to exploiting the 3D models is to pretrain the network in the traditional way (A1) and then finetune it with EPro-PnP (B5), yielding significantly better results (73.87). This training scheme partially benefits from more training epochs (2\texttimes160 in total). Furthermore, keeping the coordinate regression loss during finetuning (B6) slightly improves the score (73.95 vs. 73.87).

We also observe that both the derivative regularization loss (B2) and the coordinate regression loss (B3) improve the results of the bare KL loss setup (B0) to similar extends (67.36 vs. 67.74), as they are both disentangled objectives.

\subsection{Comparison to Implicit Differentiation and Reprojection-Based Loss}

As shown in Table~\ref{tab:losscompare}, when the coordinate regression loss is removed, i.e., object 3D models are unavailable, both implicit differentiation and reprojection loss fail to learn the pose properly. Yet EPro-PnP manages to learn the 3D coordinates and weights from scratch. This validates that EPro-PnP can be used as a general pose estimator without relying on geometric prior.

\begin{table}[!t]
\caption{Results on EPro-PnP-enhanced CDPN. $\mathcal{L}_\text{crd}$ refers to the masked coordinate regression loss in the original \cite{CDPN}, here the loss is imposed only on $x^\text{3D}$, not $w^\text{2D}$. Init. refers to initializing the model with pretrained weights from A1.}
\label{tab:cdpnablation}
\centering
    \scalebox{1.0}{%
    \setlength{\tabcolsep}{0.4em}
    \begin{tabular}{ccccccccl}
        \toprule
        \multirow{2}[2]{*}{ID} & \multirow{2}[2]{*}{$\mathcal{L}_\text{KL}$} & \multirow{2}[2]{*}{$\mathcal{L}_\text{reg}$} & \multirow{2}[2]{*}{$\mathcal{L}_\text{crd}$} & \multirow{2}[2]{*}{Init.} & \multicolumn{3}{c}{ADD(-S)} & \multirow{2}[2]{*}{Mean} \\
        \cmidrule(lr){6-8}
        {} & {} & {} & {} & {} & 0.02d & 0.05d & 0.1d \\
        \midrule
        B0 & \checkmark & {} & {} & {} & 28.48 & 67.20 & 89.93 & 61.87 \\
        B1 & {} & \checkmark & {} & {} & 25.86 & 70.90 & 92.68 & 63.15 \\
        B2 & \checkmark & \checkmark & {} & {} & 34.08 & 74.16 & 93.85 & 67.36 \\
        \midrule
        B3 & \checkmark & {} & \checkmark & {} & 34.40 & 75.00 & 93.83 & 67.74 \\
        \midrule
        B4 & \checkmark & \checkmark & \checkmark & {} & 36.22 & 75.97 & 94.64 & 68.94 \\
        B5 & \checkmark & \checkmark & {} & \checkmark & 43.34 & 82.13 & 96.14 & 73.87 \\
        B6 & \checkmark & \checkmark & \checkmark & \checkmark & 43.77 & 81.73 & 96.36 & 73.95 \\
        \bottomrule
    \end{tabular}}
\end{table}

\begin{table}[!t]
\caption{Comparison among loss functions by experiments conducted on the same dense correspondence network. For implicit differentiation, we minimize the distance metric of pose in Eq.~(\ref{regloss}) instead of the reprojection-metric pose loss in BPnP~\cite{BPnP}.}
\label{tab:losscompare}
\centering
    \scalebox{1.0}{%
    \setlength{\tabcolsep}{0.5em}
    \begin{tabular}{lcccccc}
        \toprule
        Main Loss & $\mathcal{L}_\text{crd}$ & 2\textdegree & 2 cm & 2\textdegree, 2 cm & \makecell{ADD(-S) \\ 0.1d} \\
        \midrule
        Implicit diff.~\cite{BPnP} & {} & \multicolumn{4}{c}{divergence} \\
        Reprojection~\cite{monorun} & {} & \phantom{0}0.32 & 42.30 & \phantom{0}0.16 & 14.56 \\
        KL div. (ours) & {} & 58.28 & 91.17 & 55.71 & 89.93 \\
        \midrule
        Implicit diff.~\cite{BPnP} & \checkmark & 56.13 & 91.13 & 53.33 & 88.74 \\
        Reprojection~\cite{monorun} & \checkmark & 62.79 & 92.91 & 60.65 & 92.04 \\
        KL div. (ours) & \checkmark & 69.95 & 94.97 & 68.38 & 93.83\\
        \bottomrule
    \end{tabular}}
\end{table}

\begin{table}[!t]
\caption{Comparison to the state-of-the-art geometric methods. BPnP~\cite{BPnP} is not included as it adopts a different train/test split. *Although GDRNet~\cite{gdrnet} only reports the performance in its ablation section, it is still a fair comparison to our method, since both use the same baseline (CDPN).}
\label{tab:sota}
\centering
    \scalebox{1.0}{%
    \setlength{\tabcolsep}{0.4em}
    \begin{tabular}{llccc}
        \toprule
        \multirow{2}[2]{*}{Method} & \multirow{2}[2]{*}{Type} & \multicolumn{3}{c}{ADD(-S)} \\
        \cmidrule(lr){3-5}
        {} & {} & 0.02d & 0.05d & 0.1d \\
        \midrule
        CDPN~\cite{CDPN} & PnP + Explicit depth & - & - & 89.86 \\
        HybridPose~\cite{hybridpose} & Hybrid constraints & - & - & 91.3\phantom{0} \\
        GDRNet*~\cite{gdrnet} & PnP + Explicit depth & 35.6\phantom{0} & 76.0\phantom{0} & 93.6\phantom{0} \\
        DPOD~\cite{DPOD} & PnP + Explicit refiner & - & - & 95.15 \\
        PVNet-RePOSE~\cite{repose} & PnP + Implicit refiner & - & - & 96.1\phantom{0} \\
        PVNet-RNNPose~\cite{rnnpose} & PnP + Implicit refiner & 50.39 & 85.56 & 97.37 \\
        \midrule
        Ours & PnP & 43.77 & 81.73 & 96.36 \\
        \bottomrule
    \end{tabular}}
\end{table}

\subsection{Comparison to the State of the Art}

As shown in Table~\ref{tab:sota}, although we base EPro-PnP on the older baseline CDPN~\cite{CDPN}, the results are better than some of the more advanced methods, e.g., the pose refiner RePOSE~\cite{repose} that adds extra overhead to the PnP-based initial estimator PVNet~\cite{pvnet}. Among all these entries, EPro-PnP is the most straightforward as it simply solves the PnP problem itself, without refinement network~\cite{repose, DPOD, rnnpose}, explicit depth prediction~\cite{CDPN,gdrnet}, or multiple representations~\cite{hybridpose}. 

Moreover, removing the translation head (depth prediction) from the original CDPN-Full results in far fewer parameters in our model (from 113M to 27M) , and the overall inference speed is more than twice as fast as CDPN-Full (including dataloading, measured at a batch size of 32), even though we introduce the iterative LM solver.  Furthermore, faster inference is possible if the number of points $N=64\times64$ is reduced to an optimal level.

\subsection{Visualizations}

\begin{figure*}[t]
   \begin{center}
   \includegraphics[width=0.84\textwidth]{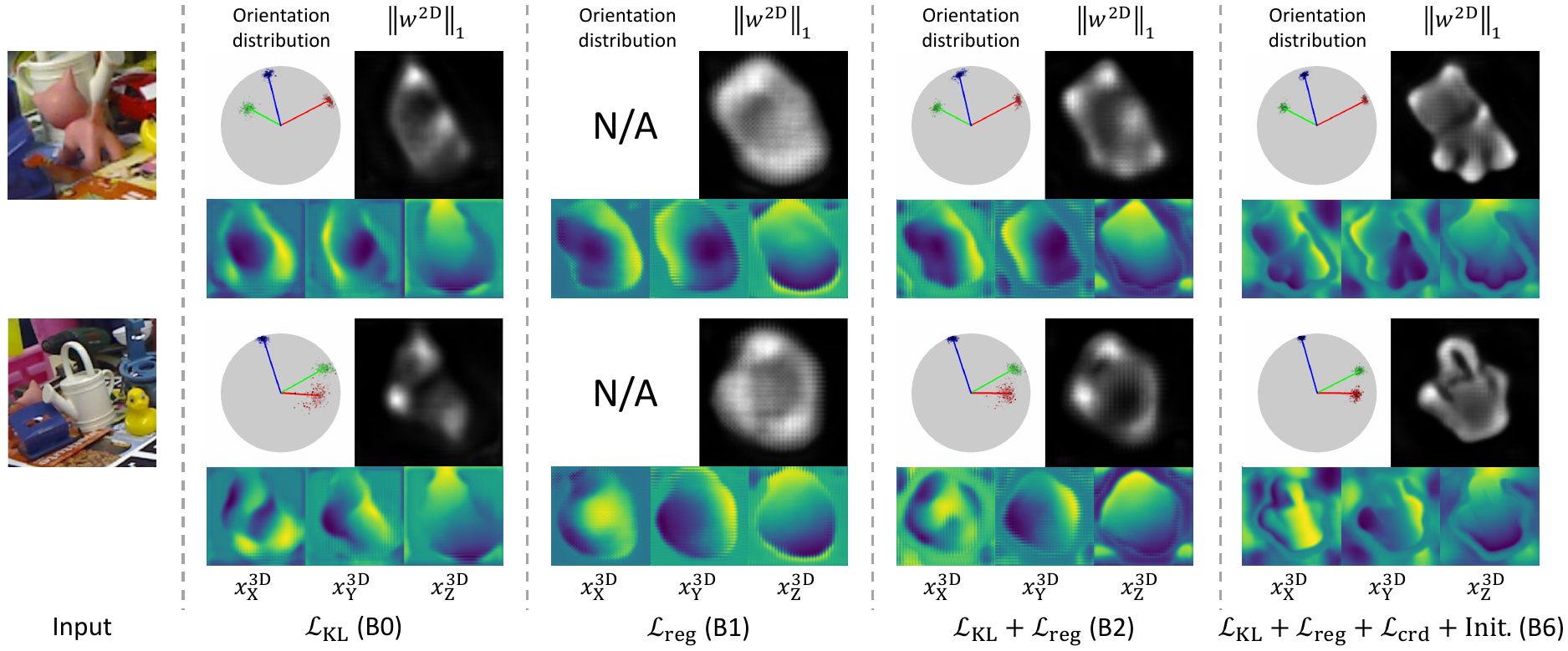}
   \end{center}
   \vspace{-0.5ex}
   \caption{Visualizations of the inferred orientation distributions, weight maps, and coordinate maps on LineMOD test set.} 
\label{fig:cdpnviz}
\end{figure*}

As illustrated in Figure~\ref{fig:cdpnviz}, the weight maps predicted by the model trained with the KL loss (B0) tend to be more focused on important parts of the objects (e.g., the head and handle of the watering can), while those with the derivative regularization loss (B1) are more evenly spread out. Combining the two loss functions (B2) leads to more reasonable weighting, and more details in the object geometry (represented by $x^\text{3D}$). With additional geometry pretraining and supervision (B6), the model outputs sharper correspondence maps, which contribute to higher pose accuracy and lower entropy of the probabilistic pose.

\section{3D Object Detection based on Deformable Correspondence Network}
To demonstrate that EPro-PnP can learn the entire set of 2D-3D correspondences $\left\{x^\text{3D}_i,x^\text{2D}_i,w^\text{2D}_i\right\}_{i=1}^N$ from scratch, and the possibility of designing novel correspondence networks capable of handling pose ambiguity, we propose a novel \emph{deformable correspondence network} for 3D object detection. The network owes its name to the Deformable DETR~\cite{deformabledetr}, a work that inspired our model architecture.

\subsection{Network Architecture}

\begin{figure*}[t]
   \begin{center}
   \includegraphics[width=0.9\textwidth]{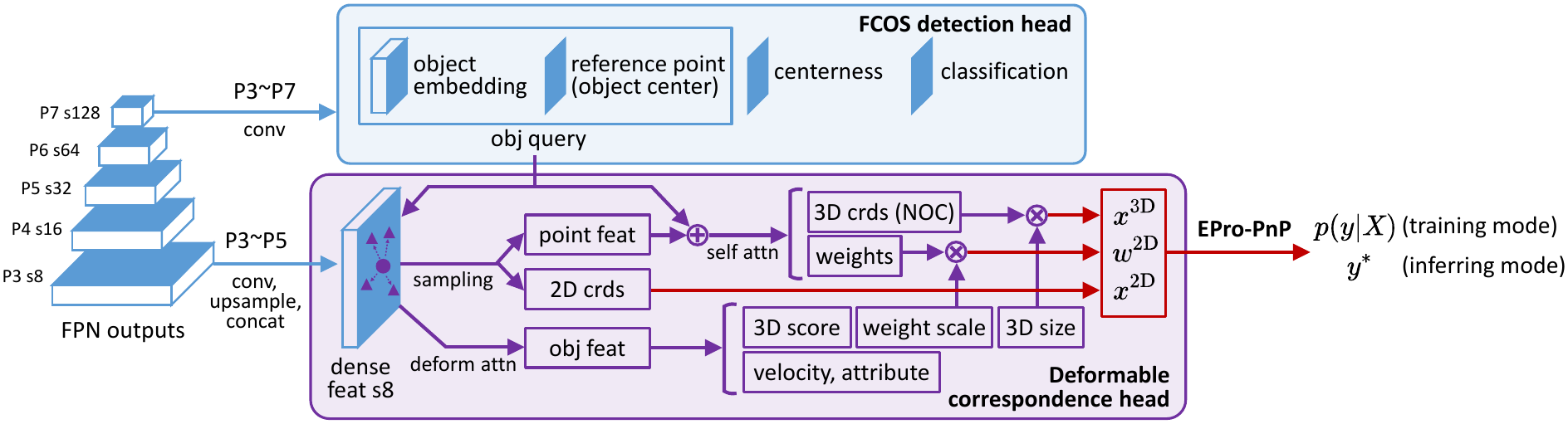}
   \end{center}
   \vspace{-0.2ex}
   \caption{The deformable correspondence network based on the FCOS3D~\cite{fcos3d} detector. Note that the sampled point-wise features are shared by the point-level subnet and the deformable attention layer that aggregates the features for object-level predictions.} 
\label{fig:fcos3d}
\end{figure*}

As shown in Figure~\ref{fig:fcos3d}, the deformable correspondence network is an extension of the FCOS3D~\cite{fcos3d} framework. The original FCOS3D is a one-stage detector that directly regresses the center offset, depth, and yaw orientation of multiple objects for 4DoF pose estimation. In our adaptation, the outputs of the multi-level FCOS head~\cite{fcos} are modified to generate object queries instead of directly predicting the pose. Inspired by Deformable DETR~\cite{deformabledetr}, the appearance and position of a query is disentangled into the object embedding vector and the reference point. Moreover, to better distinguish objects of different classes, we learn a set of class embedding vectors, one of which will be selected according to the object label to be aggregated into the object embedding vector via addition (not shown in Figure~\ref{fig:fcos3d} for brevity).

With the object queries, a multi-head deformable attention layer~\cite{deformabledetr} is adopted to sample the key-value pairs from interpolated dense feature map, with the value projected into \emph{point-wise features} (point feat), and meanwhile aggregated into the \emph{object-level features} (obj feat). 

The point features are passed into a subnet that predicts the 3D points and corresponding weights (normalized by Softmax). Following MonoRUn~\cite{monorun}, the 3D points are set in the normalized object coordinate (NOC) space to handle categorical objects of various sizes.

The object features are responsible for predicting the object-level properties: (a) the 3D score (i.e., 3D localization confidence), (b) the global weight scale, (c) the 3D box size for recovering the absolute scale of the 3D points, and (d) other optional properties (velocity, attribute) required by the nuScenes benchmark~\cite{nuscenes}.

\subsubsection{Implementation Details}
We adopt the same detector architecture as in FCOS3D~\cite{fcos3d}, with ResNet-101-DCN~\cite{dcn} as backbone. The deformable correspondence head predicts $N=128$ pairs of 2D-3D points.
The network is trained for 12 epochs by the AdamW~\cite{adamw} optimizer, with a batch size of 12 images across 2 GPUs on the nuScenes dataset~\cite{nuscenes}.

\subsection{Loss Functions}

\subsubsection{Correspondence Loss}

The deformable 2D-3D correspondences can be learned solely with the KL divergence loss $\mathcal{L}_\text{KL}$, or in conjunction with the regularization loss $\mathcal{L}_\text{reg}$.

\subsubsection{Auxiliary Correspondence Loss (Optional)}
\label{auxcorrloss}

Inspired by the dense correspondence network MonoRUn~\cite{monorun}, we regularize the dense features by appending a small auxiliary network that predicts the multi-head dense 3D coordinates and weights corresponding to densely-sampled 2D points within the ground truth (using RoI Align~\cite{maskrcnn}). This allows us to employ the uncertainty-aware reprojection loss $\mathcal{L}_\text{proj}$~\cite{monorun} without directly regularizing the deformable correspondences. Furthermore, we can convert the LiDAR scan of objects into sparse 3D object coordinate maps, so that the classical coordinate regression loss $\mathcal{L}_\text{crd}$ can be imposed on the auxiliary branch as well. Both of the loss functions are implemented as the NLL of Gaussian mixtures to deal with ambiguity (see supplementary for details).

\subsubsection{Other Loss Functions}

Loss functions on the FCOS head include:
\begin{itemize}
    \item Basic detector loss, including focal loss~\cite{focalloss} for classification and cross entropy loss for centerness.
    \item A smooth L1 loss for regressing the 2D reference points, with the target defined as the center of the visible region of the objects.
    \item A GIoU loss~\cite{giou} for auxiliary 2D box regression, following the 2D auxiliary supervision in M\textsuperscript{2}BEV~\cite{m2bev}.
\end{itemize}

Loss functions for the object-level predictions include:
\begin{itemize}
    \item A cross entropy loss for the 3D score.
    \item A smooth L1 loss for regressing the 3D box size.
    \item A smooth L1 loss for regressing the velocity and a cross entropy loss for attribute classification.
\end{itemize}

Additionally, inspired by DD3D~\cite{dd3d}, we further exploit the available LiDAR data to build an auxiliary depth supervision. By projecting the LiDAR points to the camera frame, we extract the point-wise features from the interpolated dense feature map, which are then fed into a small 2-layer MLP to predict the scene depth. Same as the auxiliary correspondence loss functions in Section~\ref{auxcorrloss}, the depth loss is implemented as the NLL of Gaussian mixtures, which allows modeling discontinuities around sharp edges~\cite{smdnets}.

\subsection{Dataset and Metrics}

We evaluate the deformable correspondence network on the nuScenes 3D object detection benchmark~\cite{nuscenes}, which provides a large scale of data collected in 1000 scenes.
Each scene contains 40 keyframes, annotated with a total of 1.4M 3D bounding boxes from 10 categories.
Each keyframe includes 6 RGB images collected from surrounding cameras. The data is split into 700/150/150 scenes for training/validation/testing. The official benchmark evaluates the average precision with true positives judged by 2D center error on the ground plane.
The mAP metric is computed by averaging over the thresholds of 0.5, 1, 2, 4 meters. Besides, there are 5 true positive metrics: Average Translation Error (ATE),
Average Scale Error (ASE),
Average Orientation Error (AOE),
Average Velocity Error (AVE)
and Average Attribute Error (AAE).
Finally, there is a nuScenes detection score (NDS) computed as a weighted average of the above metrics. 

\subsection{Main Results and Discussions}

\subsubsection{Comparison Among Correspondence Loss Functions}

As shown in Table~\ref{tab:deformloss}, the model trained with KL loss alone (C0) is significantly stronger than the model trained with the derivative regularization loss alone (C1) in all the metrics of concern, especially the orientation error (0.332 vs. 0.607). This is due to the presence of orientation ambiguity in the nuScenes dataset. Even if all the auxiliary loss functions (C2) are applied, the derivative regularization loss still fail to reach comparable performance to the Monte Carlo KL loss. Adding up all the loss functions (C3), the results can be boosted even further.

\begin{table}[!t]
\caption{Experiments on the nuScenes validation set.}
\label{tab:deformloss}
\centering
    \scalebox{1.0}{%
    \setlength{\tabcolsep}{0.4em}
    \begin{tabular}{cccccccccc}
        \toprule
        \multirow{2}[2]{*}{ID} & \multirow{2}[2]{*}{$\mathcal{L}_\text{KL}$} & \multirow{2}[2]{*}{$\mathcal{L}_\text{reg}$} & \multicolumn{2}{c}{Aux. loss} & \multirow{2}[2]{*}{NDS\textuparrow} & \multirow{2}[2]{*}{mAP\textuparrow} & \multirow{2}[2]{*}{mATE\textdownarrow} & \multirow{2}[2]{*}{mAOE\textdownarrow} \\
        \cmidrule(lr){4-5}
        {} & {} & {} & $\mathcal{L}_\text{crd}$ & $\mathcal{L}_\text{proj}$ \\
        \midrule
        C0 & \checkmark &  {} & {} & {} & 0.447 & 0.380 & 0.656 & 0.332 \\
        C1 & {} & \checkmark &  {} & {} & 0.408 & 0.363 & 0.683 & 0.607 \\
        C2 & {} & \checkmark & \checkmark & \checkmark &  0.429 & 0.363 & 0.691 & 0.397 \\
        C3 & \checkmark & \checkmark & \checkmark & \checkmark & 0.463 & 0.392 & 0.626 & 0.282 \\
        \bottomrule
    \end{tabular}}
\end{table}

\begin{table*}[t]
\caption{Comparison to the state-of-the-art single-frame image-based 3D object detectors on the nuScenes test set. Methods with extra pretraining other than ImageNet backbone are not included for comparison. \S\ indicates test-time flip augmentation (TTA). \textdagger\ indicates model ensemble.}
\label{tab:nuscenestest}
\centering
    \scalebox{1.0}{%
    \setlength{\tabcolsep}{0.5em}
    \begin{tabular}{llcccccccc}
        \toprule
        Method & Backbone & NDS\textuparrow & mAP\textuparrow & mATE\textdownarrow & mASE\textdownarrow & mAOE\textdownarrow & mAVE\textdownarrow & mAAE\textdownarrow \\
        \midrule
        FCOS3D \S\textdagger~\cite{fcos3d} & R101 & 0.428 & 0.358 & 0.690 & 0.249 & 0.452 & 1.434 & 0.124 \\
        PGD \S~\cite{pgd} & R101 & 0.448 & 0.386 & 0.626 & 0.245 & 0.451 & 1.509 & 0.127 \\
        PETR~\cite{petr} & R101 & 0.455 & 0.391 & 0.647 & 0.251 & 0.433 & 0.933 & 0.143 \\
        BEVFormer~\cite{bevformer} & R101 & 0.462 & 0.409 & 0.650 & 0.261 & 0.439 & 0.925 & 0.147 \\
        PolarFormer~\cite{polarformer} & R101 & 0.470 & 0.415 & 0.657 & 0.263 & 0.405 & \textbf{0.911} & 0.139 \\
        PETR~\cite{petr} & Swin-B & 0.483 & \textbf{0.445} & 0.627 & 0.249 & 0.449 & 0.927 & 0.141 \\
        \midrule
        Ours & R101 & 0.481 & 0.409 & 0.559 & 0.239 & 0.325 & 1.090 & \textbf{0.115} \\
        Ours \S & R101 & \textbf{0.490} & 0.423 & \textbf{0.547} & \textbf{0.236} & \textbf{0.302} & 1.071 & 0.123 \\
        \bottomrule
    \end{tabular}}
\end{table*}

\subsubsection{Comparison to the State of the Art}

Results on the nuScenes test set~\cite{nuscenes} are shown in Table~\ref{tab:nuscenestest}. At the time of submitting the manuscript (Jan 2023), EPro-PnP is the No.~1 single-frame monocular 3D object detector without extra data, according to the official nuScenes detection leaderboard. Among the models using ResNet-101 as backbones, EPro-PnP outperforms PolarFormer~\cite{polarformer} by a clear margin (NDS 0.481 vs. 0.470), despite basing the deformable correspondence network on the older FCOS detector. With test-time flip augmentation (following FCOS3D~\cite{fcos3d}), our model even outperforms PGD~\cite{pgd} with the bulky Swin-B~\cite{Swin} backbone.

Since EPro-PnP is targeted at improving pose accuracy, it is not surprising to see that our model obtains exceptional results regarding the mATE and mAOE metrics, outperforming PolarFormer by a wide margin (mATE 0.559 vs. 0.657, mAOE 0.325 vs. 0.405).

It is worth noting that, EPro-PnP is currently the only method among the entries in Table~\ref{tab:nuscenestest} that utilizes geometric pose reasoning, which is not a popular choice because previous non-end-to-end geometric methods usually fall behind when trained on large-scale real-life data.

\subsection{Visualizations}

\begin{figure}[!t]
\centering
\includegraphics[width=0.9\linewidth]{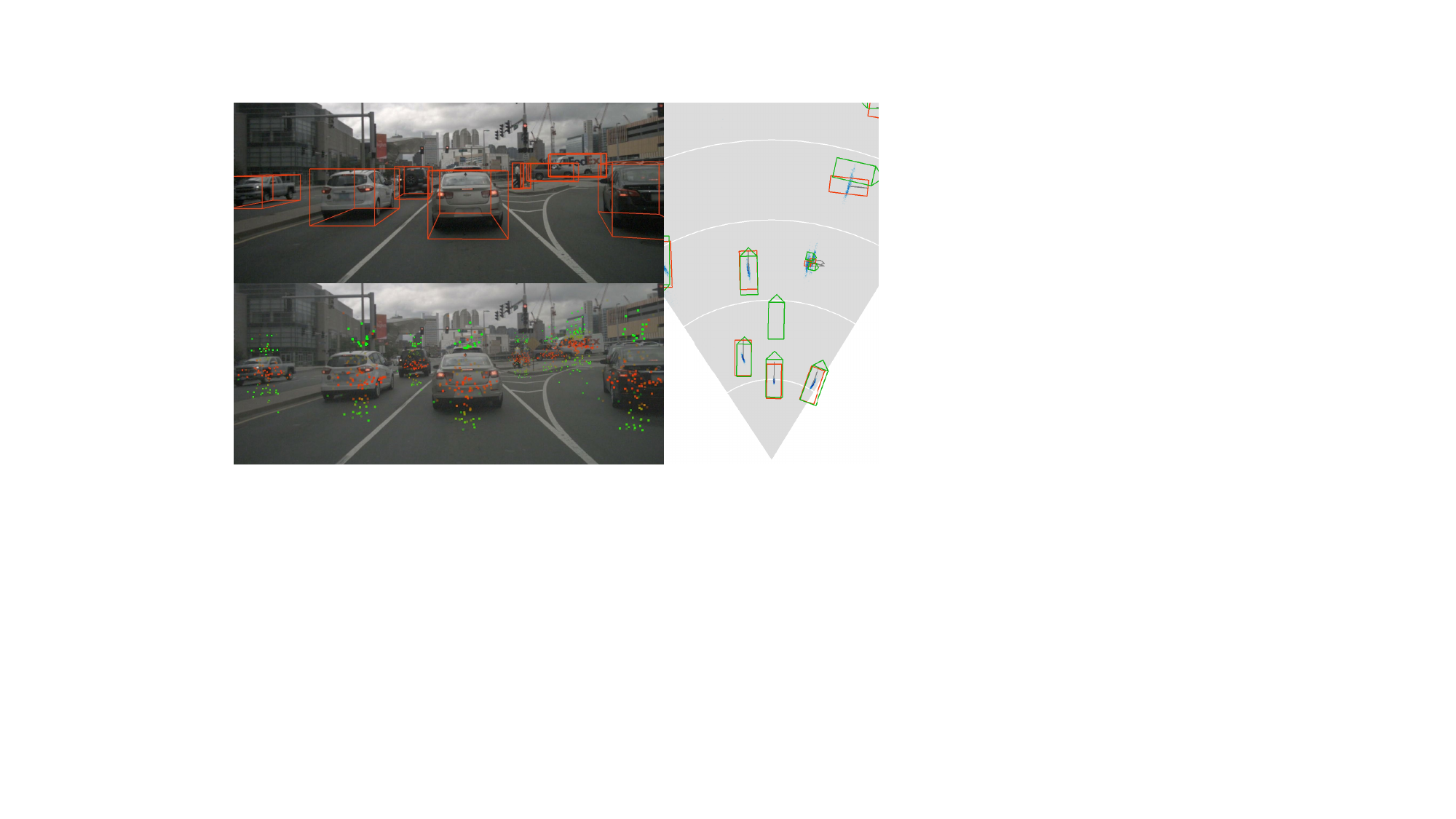}
\caption{Inferred results on nuScenes validation set. On the top-left are the predicted 3D bounding boxes. On the bottom-left are the 2D points $x^\text{2D}$ colored by the XY corresponding weights $w^\text{2D}$ (red indicates $w^\text{2D}_\text{X} > w^\text{2D}_\text{Y}$, green indicates $w^\text{2D}_\text{X} < w^\text{2D}_\text{Y}$). On the right side are the inferred bounding boxes (red), marginal position density (blue), marginal orientation density (grey line), and ground truth bounding boxes (green) in bird's eye view.}
\label{fig:nusviz}
\end{figure}

An example of the monocular detection result is shown in Figure~\ref{fig:nusviz}. We observe that the red 2D points (indicating greater $x^\text{3D}$ in the X axis) are usually spread right over the objects, which mainly determines the orientation, while the green 2D points (indicating greater $x^\text{3D}$ in the Y axis) are off the top and bottom of the objects, which determines the position (mainly the depth). It seems that the network learns to associate object depth to the height of the object's projection, since the height invariant to 1D orientation in the ground plane.

\begin{figure}[!t]
\centering
\includegraphics[width=0.8\linewidth]{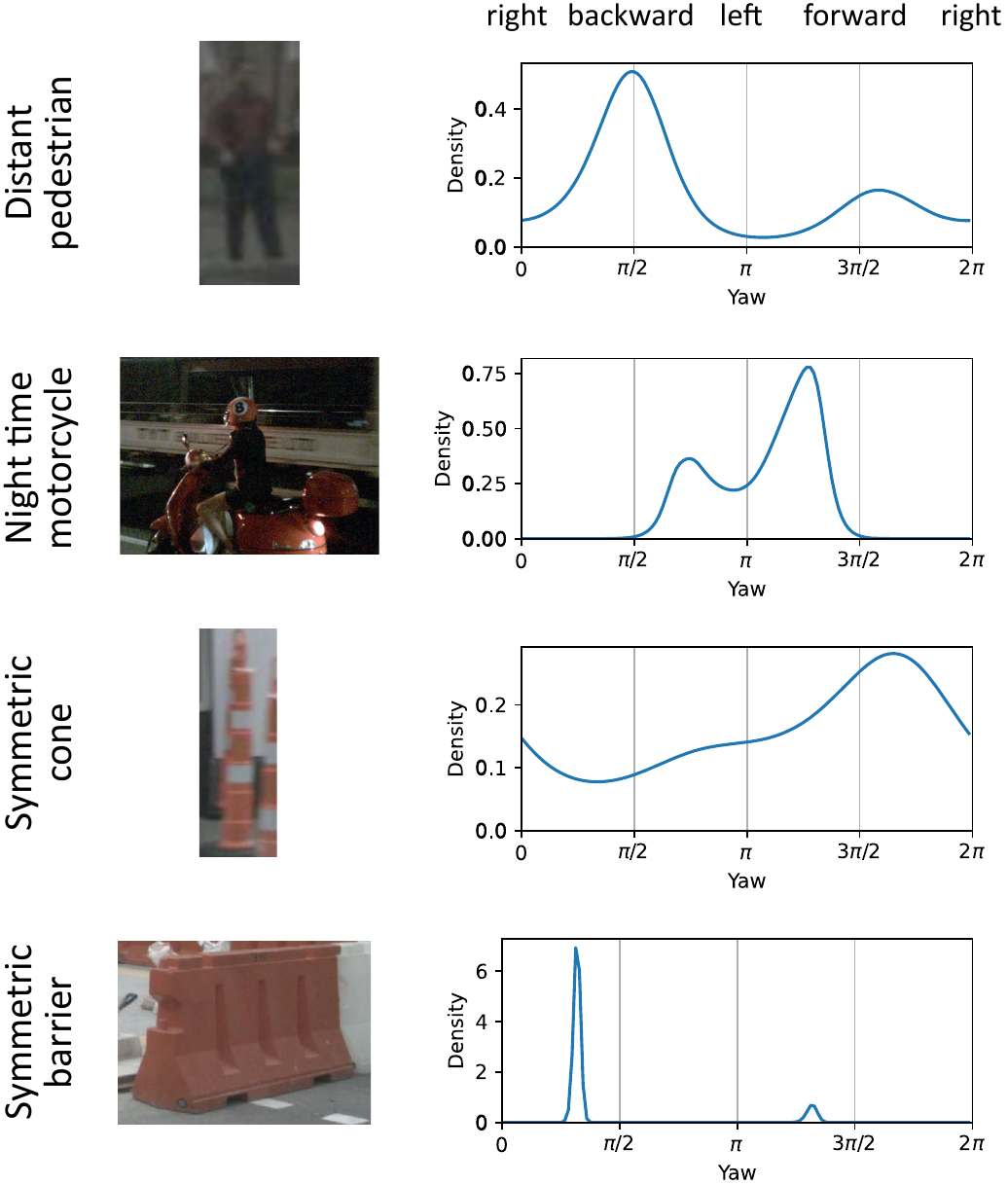}
\caption{Examples of ambiguous orientations in the nuScenes dataset and the predicted orientation distribution (conditioned on the optimal position). The ambiguity originates from either uncertain observations (e.g., distant pedestrian, night time motorcycle), or symmetric objects.}
\label{fig:ambiguity}
\end{figure}

Figure~\ref{fig:ambiguity} shows that the flexibility of EPro-PnP allows predicting multimodal distributions with strong expressive power, successfully capturing the orientation ambiguity without discrete multi-bin classification~\cite{fcos3d,mousavian20173d} or complicated mixture model~\cite{bui20206d}.

\subsection{Inference Time}

The average inference time per frame (comprising a batch of 6 surrounding 1600\texttimes672 images, without TTA) is shown in Table~\ref{tab:runtime}, measured on RTX 3090 GPU and Core i9-10920X CPU. On average, the batch PnP solver takes 26 ms/46 ms processing 655.3 objects per frame, before non-maximum suppression (NMS).

\begin{table}[h]
\caption{Inference time (sec) of the deformable correspondence network on nuScenes~\cite{nuscenes}. The PnP solver (including initialization) works faster (26 ms) with PyTorch v1.8.1, for which the code was originally developed, while the full model works faster (304 ms) with PyTorch v1.10.1.}
\label{tab:runtime}
\centering
    \scalebox{1.0}{%
    \setlength{\tabcolsep}{0.0em}
    \begin{tabular}{@{\hskip 0.28cm}p{2cm} >{\centering\arraybackslash}p{1.2cm} >{\centering\arraybackslash}p{1.2cm} >{\centering\arraybackslash}p{1.2cm} >{\centering\arraybackslash}p{1.2cm} >{\centering\arraybackslash}p{1.2cm}}
        \toprule
        \multirow{2}[2]{*}{PyTorch} & \multirow{2}[2]{*}{\makecell{Backbone \\ \& FPN}} & \multicolumn{2}{c}{Heads} & 
        \multirow{2}[2]{*}{PnP} & 
        \multirow{2}[2]{*}{Total} \\
        \cmidrule(lr){3-4}
        {} & {} & FCOS & Deform \\
        \midrule
        v1.8.1+cu111 & 0.194 & 0.074 & 0.029 & \textbf{0.026} & 0.328 \\
        v1.10.1+cu113 & 0.173 & 0.056 & 0.025 & 0.046 & \textbf{0.304} \\
        \bottomrule
    \end{tabular}}
\end{table}

\section{Limitations}

Training the network with the Monte Carlo pose loss is inevitably \emph{slower} than the baseline. With the batch size of 32 on a GTX 1080 Ti GPU, training the CDPN (without translation head) takes 143 seconds per epoch with the original coordinate regression loss, and 241 seconds per epoch with the Monte Carlo pose loss, which is about 70\% longer time. However, the training time can be controlled by adjusting the number of Monte Carlo samples or the number of 2D-3D corresponding points. 

Although the underlying principles are theoretically generalizable to other learning models with nested optimization layer, known as declarative networks~\cite{declarative}, the Monte Carlo pose loss would become impractical with the growth of dimensionality.

While EPro-PnP seems to be a universal approach to end-to-end geometric pose estimation, it should be noted that the design of 2D-3D correspondence network still plays a major role in the model. For example, simply removing the 2D box size from Figure~\ref{fig:cdpnnet} would result in a notable decrease in pose accuracy. Future work may explore the feature-metric correspondence in \cite{repose, rnnpose, monojsg} as a more expressive alternative to plain Euclidean reprojection error.

\section{Conclusion}

This paper proposes the EPro-PnP, which translates the non-differentiable PnP operation into a differentiable probabilistic layer, empowering end-to-end 2D-3D correspondence learning of unprecedented flexibility. The connections to previous work~\cite{monorun,BPnP,blindpnp,dsac++,repose} have been thoroughly discussed with theoretical and experimental proofs, revealing the contributions of the Monte Carlo KL loss and the derivative regularization loss. For application, EPro-PnP can be simply integrated into existing PnP-based networks, or inspire novel solutions such as the deformable correspondence network.

\ifCLASSOPTIONcompsoc
  \section*{Acknowledgments}
\else
  \section*{Acknowledgment}
\fi

This project was supported by the National Natural Science Foundation of China [No.~52002285], the Shanghai Science and Technology Commission [No.~21ZR1467400], the original research project of Tongji University [No.~22120220593], and the National Key R\&D Program of China [No.~2021YFB2501104]. Part of the work was done when H. Chen was interning at Alibaba Group, supported by the Alibaba Research Intern Program.

\ifCLASSOPTIONcaptionsoff
  \newpage
\fi



%\bibliographystyle{IEEEtran}

\bibliographystyle{IEEEtran}
\bibliography{egbib}


% H.~Kopka and P.~W. Daly, \emph{A Guide to {\LaTeX}}, 3rd~ed.\hskip 1em plus

% \end{thebibliography}


\vfill

\begin{IEEEbiography}[{\includegraphics[width=1in,height=1.25in,clip,keepaspectratio]{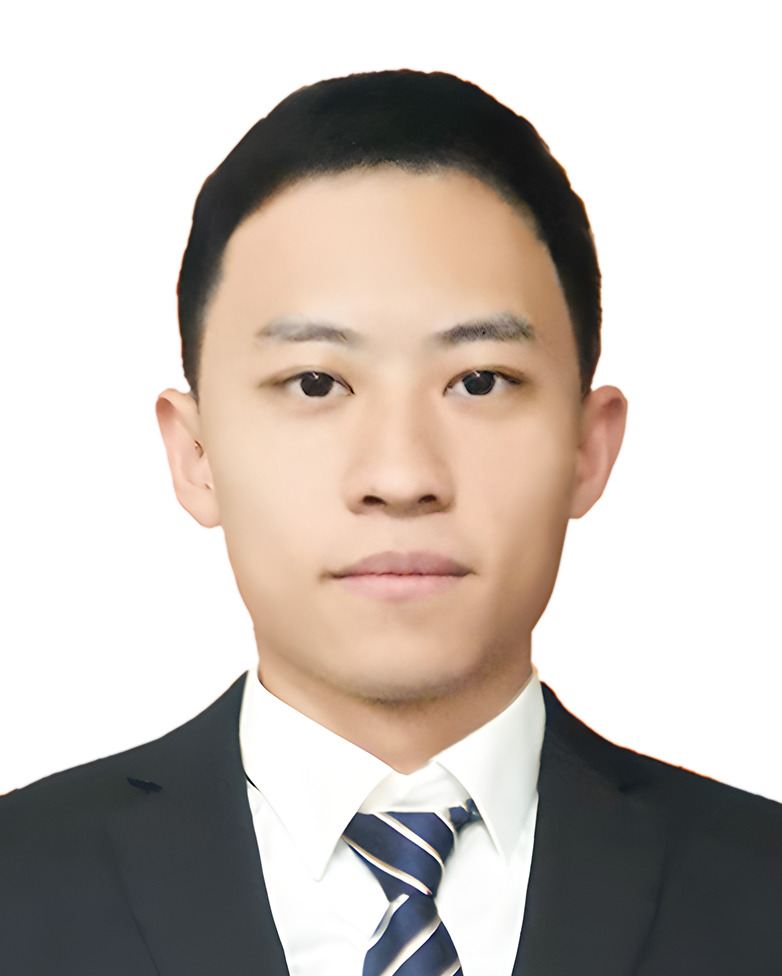}}]{Hansheng Chen} received the M.S.E. and B.E. degree in vehicle engineering from Tongji University, Shanghai, in 2020 and 2023, respectively. He is pursuing the PhD degree in the Computer Science Department of Stanford University. His current research interests lie in computer graphics and 3D vision, with a specific focus on 3D generation, reconstruction, editing, and neural rendering.
\end{IEEEbiography}

\begin{IEEEbiography}[{\includegraphics[width=1in, height=1.25in, trim = 20mm 30mm 20mm 0mm, clip=true,keepaspectratio]{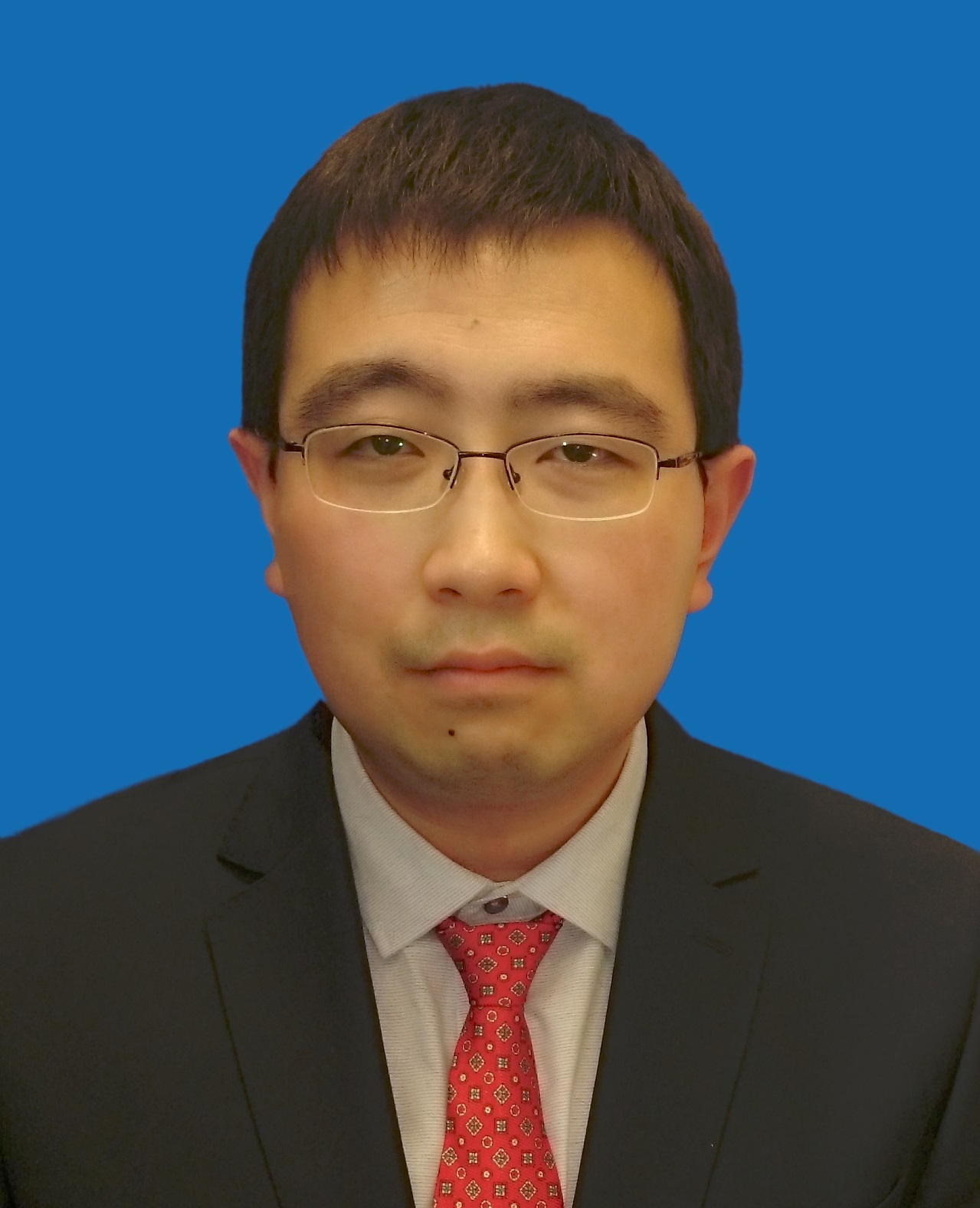}}]{Wei Tian} received the B.Sc degree in mechatronics engineering from Tongji University, Shanghai, China, in 2010, and received his the M.Sc. degree in electrical engineering and information technology at KIT, Karlsruhe, Germany, in May 2013. From October 2013, he was with the institute of measurement and control systems at KIT and received the Ph.D. degree in January 2019. 
%Afterwards, he continued his post-doctoral research at KIT. 
From May 2020, he is a leader of comprehensive perception research group at School of Automotive Studies, Tongji University. He is currently working on research areas of robust object detection and trajectory prediction.
\end{IEEEbiography}

\begin{IEEEbiography}[{\includegraphics[width=1in,height=1.25in,clip,keepaspectratio]{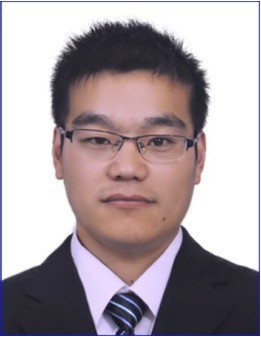}}]{Pichao Wang} received the Ph.D. degree in computer science from the University of Wollongong, Wollongong, NSW, Australia. He is currently a Senior Research Scientist in Amazon Prime Video, USA. He has authored 90+ peer reviewed papers, including those in highly regarded journals and conferences such as IJCV, IEEE TMM, CVPR, ICCV, ECCV, ICLR, AAAI, ACM MM, etc. He is the recipient of CVPR2022 Best Student Paper Award. He is named AI 2000 Most Influential Scholar during 2012-2022 by Miner, due to his contributions in the field of multimedia. He is also in the list of World’s Top 2\% Scientists named by Stanford University. He serves as the Area Chair of ICME 2021,2022. He also serves as an Associate Editor of Journal of Computer Science and Technology (Tier 1, CCF B).
\end{IEEEbiography}

\begin{IEEEbiography}[{\includegraphics[width=1in,height=1.25in,trim = 35mm 0mm 35mm 0mm,clip,keepaspectratio]{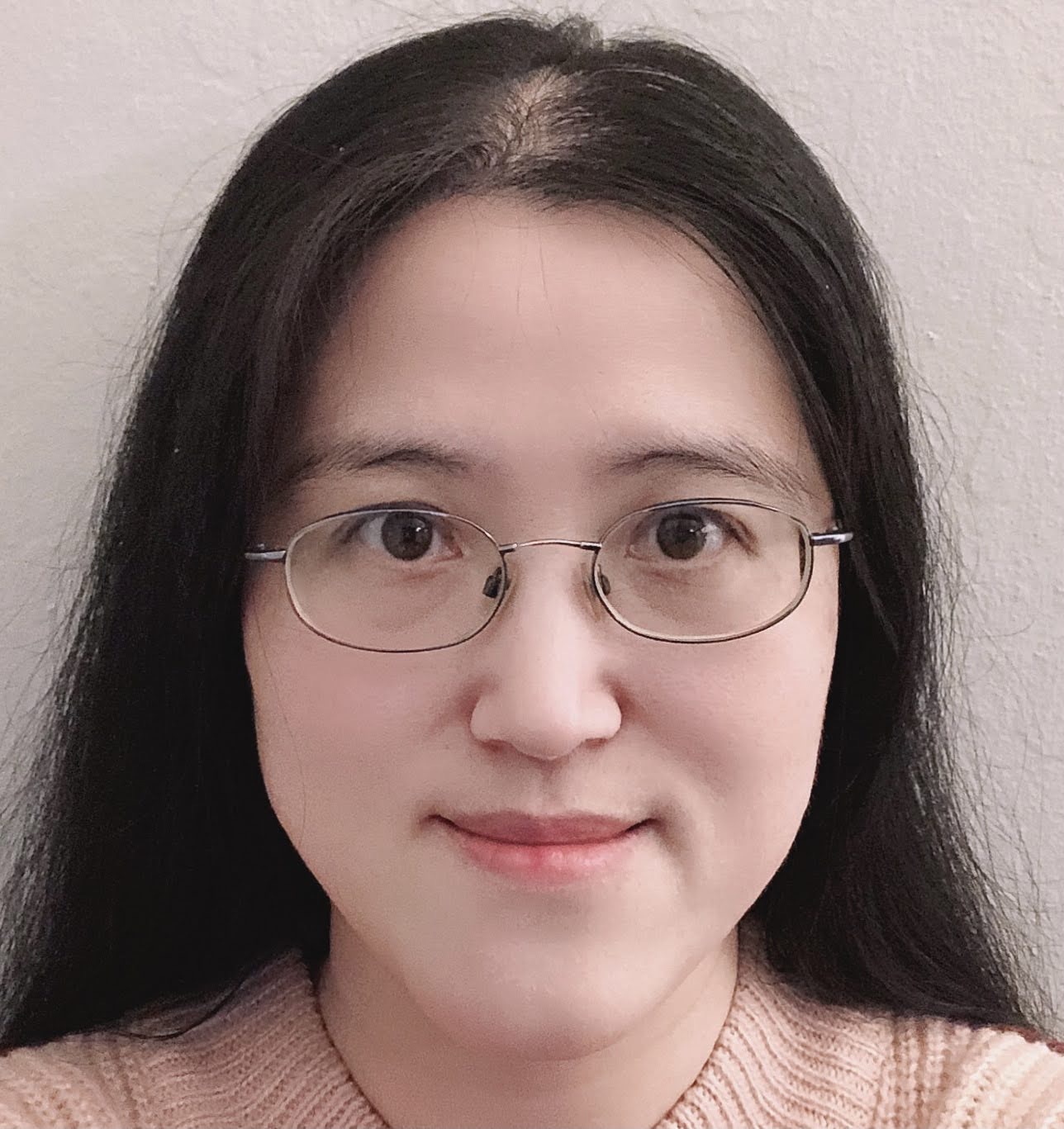}}]{Fan Wang} received the B.S. and M.S. degrees from the Department of Automation, Tsinghua University, Beijing, China, and the Ph.D. degree from the Department of Electrical Engineering, Stanford University, Stanford, CA, USA. She is currently working as a Senior Staff Algorithm Engineer with Alibaba Group. Her research interests include object tracking and recognition, 3D vision, and multi-sensor fusion.
\end{IEEEbiography}

\begin{IEEEbiography}[{\includegraphics[width=1in,height=1.25in,trim = 2mm 20mm 2mm 0mm,clip,keepaspectratio]{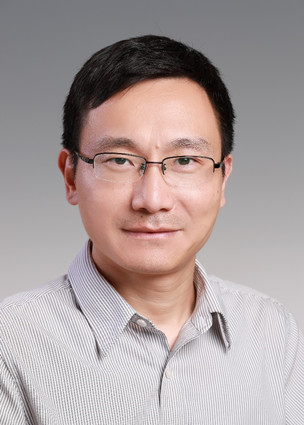}}]{Lu Xiong} received the B.E., M.E., and the Ph.D. degrees in vehicle engineering from the School of Automotive Studies, Tongji University, Shanghai, China, in 1999, 2002, and 2005, respectively. From November 2008 to 2009, he was a Postdoctoral Fellow with the Institute of Automobile Engineering and Vehicle Engines, University of Stuttgart, Stuttgart, Germany, with Dr. Jochen Wiedemann. He is currently a Professor with Tongji University and the Deputy Dean of School of Automotive Studies. His research interests include perception, decision and planning, dynamics control and state estimation, and testing and evaluation of autonomous vehicles.
\end{IEEEbiography}

\begin{IEEEbiography}[{\includegraphics[width=1in,height=1.25in,trim = 2mm 20mm 2mm 0mm,clip,keepaspectratio]{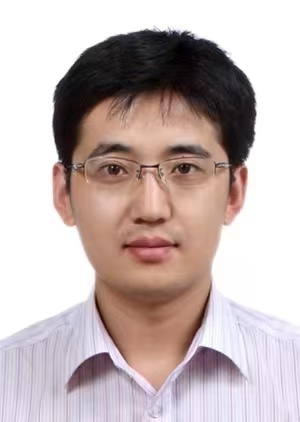}}]{Hao Li} received the Ph.D. degree from the Chinese Academy of Sciences. He is in charge of real-scene visual understanding technologies. He has published more than 20 papers and owns more than 20 licensed patents. His research interests include smart interpretation of remote sensing images, facial recognition-based clocking in systems, new retail, smart campuses, deep learning model compression, facial recognition, person re-identification, and image search.
\end{IEEEbiography}


% \vfill


\clearpage

\appendices

\section{Levenberg-Marquardt PnP Solver}

For parallel processing on GPU, we have implemented a PyTorch-based batch Levenberg-Marquardt (LM) PnP solver. The implementation generally follows the Ceres solver~\cite{ceres-solver}. Here, we discuss some important details that are related to the proposed Monte Carlo pose sampling and derivative regularization.

\subsection{LM Step with Huber Kernel}

Adding the Huber kernel influences every related aspect from the likelihood function to the LM iteration step and derivative regularization loss. Thanks to PyTorch's automatic differentiation, the robustified Monte Carlo KL divergence loss does not require much special handling. For the LM solver, however, the residual $F(y)$ (concatenated weighted reprojection errors) and the Jacobian matrix $J$ have to be rescaled before computing the robustified LM step~\cite{triggsba}.

The rescaled residual block $\tilde{f}_i(y)$ and Jacobian block $\tilde{J}_i(y)$ of the $i$-th point pair are defined as:
\begin{equation}
    \tilde{f}_i(y) = \sqrt{\rho^\prime_i} f_i(y),
\end{equation}
\begin{equation}
    \tilde{J}_i(y) = \sqrt{\rho^\prime_i} J_i(y),
\end{equation}
where
\begin{equation}
 \rho^\prime_i = 
 \begin{dcases}
 1, & \| f_i(y) \| \leq \delta,\\
 \frac{\delta}{\| f_i(y) \|}, & \| f_i(y) \| > \delta,
 \end{dcases}
\label{eqn:huber2}
\end{equation}
\begin{equation}
    J_i(y) = \frac{\partial{f_i(y)}}{\partial{y^\text{T}}}.
\end{equation}
Following the implementation of Ceres solver~\cite{ceres-solver}, the robustified LM iteration step is:
\begin{equation}
    \Delta y = -\left(\tilde{J}^\text{T}\tilde{J} + \lambda D^2 \right)^{-1} \tilde{J}^\text{T} \tilde{F},
    \label{eqn:robustlmstep}
\end{equation}
where
\begin{equation}
    \tilde{J} = 
    \begin{bmatrix}
        \tilde{J}_1(y) \\
        \vdots\\
        \tilde{J}_N(y)
    \end{bmatrix},
        \tilde{F} = 
    \begin{bmatrix}
        \tilde{f}_1(y) \\
        \vdots\\
        \tilde{f}_N(y)
    \end{bmatrix},
\end{equation}
$D$ is the square root of the diagonal of the matrix $\tilde{J}^\text{T}\tilde{J}$, and $\lambda$ is the reciprocal of the LM trust region radius~\cite{ceres-solver}.
% , and $H^\text{LM}$ compactly denotes the LM Hessian Matrix.

Note that the rescaled residual and Jacobian affects the derivative regularization, as well as the covariance estimation in the next subsection.

\subsubsection{Fast Inference Mode} 

We empirically observe that in a well-trained model, the LM trust region radius can be initialized with a very large value, effectively rendering the LM algorithm redundant. We therefore use the simple Gauss-Newton implementation for fast inference:
\begin{equation}
    \Delta y = -\left(\tilde{J}^\text{T}\tilde{J} + \varepsilon I \right)^{-1} \tilde{J}^\text{T} \tilde{F},
\end{equation}
where $\varepsilon$ is a small value for numerical stability.

\subsection{Covariance Estimation}
During training, the concentration of the AMIS proposal is determined by the local estimation of pose covariance matrix $\Sigma_{y^\ast}$, defined as:
\begin{equation}
    \Sigma_{y^\ast} = \left( \tilde{J}^\text{T} \tilde{J} + \varepsilon I \right)^{-1} \Big\rvert_{y=y^\ast},
\end{equation}
where $y^\ast$ is the LM solution that determines the location of the proposal distribution.

\section{Details on Monte Carlo Pose Sampling}

\subsection{Proposal Distribution for Position}

For the proposal distribution of the translation vector $t \in \mathbb{R}^3$, we adopt the multivariate t-distribution, with the following probability density function (PDF):
\begin{equation}
    q_\text{T}(t) = \frac{\Gamma\left(\frac{\nu + 3}{2}\right)}{\Gamma\left(\frac{\nu}{2}\right) \sqrt{\nu^3 \pi^3 |\Sigma|}} \left( 1 + \frac{1}{\nu} \| t-\mu \|_{\Sigma}^2 \right)^{\negmedspace -\frac{\nu+3}{2}},
\end{equation}
where $\| t-\mu \|_{\Sigma}^2 = (t-\mu)^\text{T} \Sigma^{-1} (t-\mu)$, with the location $\mu$, the 3\texttimes3 positive definite scale matrix $\Sigma$, and the degrees of freedom $\nu$. Following~\cite{amis}, we set $\nu$ to 3. Compared to the multivariate normal distribution, the t-distribution has a heavier tail, which is ideal for robust sampling. 

The multivariate t-distribution has been implemented in the Pyro~\cite{pyro} package.

\subsubsection{Initial Parameters}
The initial location and scale is determined by the PnP solution and covariance matrix, i.e., $\mu \gets t^\ast, \Sigma \gets \Sigma_{t^\ast}$, where $\Sigma_{t^\ast}$ is the 3\texttimes 3 submatrix of the full pose covariance $\Sigma_{p^\ast}$. Note that the actual covariance of the t-distribution is thus $\frac{\nu}{\nu-1}\Sigma_{t^\ast}$, which is intentionally scaled up for robust sampling in a wider range. 

\subsubsection{Parameter Estimation from Weighted Samples}
To update the proposal, we let the location $\mu$ and scale $\Sigma$ be the first and second moment of the weighted samples (i.e., weighted mean and covariance), respectively.

\subsection{Proposal Distribution for 1D Orientation}
For the proposal distribution of the 1D yaw-only orientation $\theta$, we adopt a mixture of von Mises and uniform distribution. The von Mises is also known as the circular normal distribution, and its PDF is given by:
\begin{equation}
    q_\text{VM}(\theta) = \frac{\exp{(\kappa \cos{(\theta - \mu)})}}{2 \pi I_0(\kappa)},
\end{equation}
where $\mu$ is the location parameter, $\kappa$ is the concentration parameter, and $I_0(\cdot)$ is the modified Bessel function with order zero. The mixture PDF is thus:
\begin{equation}
    q_\text{mix}(\theta) = (1-\alpha)q_\text{VM}(\theta) + \alpha q_\text{uniform}(\theta),
\end{equation}
with the uniform mixture weight $\alpha$. The uniform component is added in order to capture other potential modes under orientation ambiguity. We set $\alpha$ to a fixed value of $1/4$.

PyTorch has already implemented the von Mises distribution, but its random sample generation is rather slow. As an alternative we use the NumPy implementation for random sampling.

\subsubsection{Initial Parameters}
With the yaw angle $\theta^\ast$ and its variance $\sigma^2_{\theta^\ast}$ from the PnP solver, the parameters of the von Mises proposal is initialized by $\mu \gets \theta^\ast, \kappa \gets \frac{1}{3\sigma^2_{\theta^\ast}}$.

\subsubsection{Parameter Estimation from Weighted Samples}
For the location $\mu$, we simply adopt its maximum likelihood estimation, i.e., the circular mean of the weighted samples. For the concentration $\kappa$, we first compute an approximated estimation~\cite{Dhillon2003ModelingDU} by:
\begin{equation}
    \hat{\kappa} = \frac{\bar{r}(2-\bar{r}^2)}{1-\bar{r}^2},
\end{equation}
where $\bar{r} = \left\lVert \sum_j v_j[\sin{\theta_j}, \cos{\theta_j}]^\text{T} / \sum_j v_j \right\rVert$ is the norm of the mean orientation vector, with the importance weight $v_j$ for the $j$-th sample $\theta_j$. Finally, the concentration is scaled down for robust sampling, such that $\kappa \gets \hat{\kappa}/3$.

\subsection{Proposal Distribution for 3D Orientation} Regarding the quaternion-based parameterization of 3D orientation, which can be represented by a unit 4D vector $l$, we adopt the angular central Gaussian (ACG) distribution as the proposal. The support of the 4-dimensional ACG distribution is the unit hypersphere, and the PDF is given by:
\begin{equation}
    q_\text{ACG}(l) = \frac{(l^\text{T} \Lambda^{-1} l)^{-2}}{S_4 |\Lambda|^{\frac{1}{2}}},
\end{equation}
where $S_4 = 2 \pi^2$ is the 3D surface area of the 4D sphere, and $\Lambda$ is a 4\texttimes 4 positive definite matrix.

The ACG density can be derived by integrating the zero-mean multivariate normal distribution $\mathcal{N}(0, \Lambda)$ along the radial direction from $0$ to $\inf$. Therefore, drawing samples from the ACG distribution is equivalent to sampling from $\mathcal{N}(0, \Lambda)$ and then normalizing the samples to unit radius.

\subsubsection{Initial Parameters}
Consider $l^\ast$ to be the PnP solution and $\Sigma_{l^\ast}^{-1}$ to be the estimated 4\texttimes 4 inverse covariance matrix. Note that $\Sigma_{l^\ast}^{-1}$ is only valid in the local tangent space with rank 3, satisfying ${l^\ast}^\text{T} \Sigma_{l^\ast}^{-1} l^\ast = 0$. The initial parameters are determined by:
\begin{equation}
    \Lambda \gets \hat{\Lambda} + \alpha |\hat{\Lambda}|^{\frac{1}{4}}I,
    \label{acgparam}
\end{equation}
where $\hat{\Lambda} = \left(\Sigma_{l^\ast}^{-1} + I\right)^{-1}$, and $\alpha$ is a hyperparameter that controls the dispersion of the proposal for robust sampling. We set $\alpha$ to 0.001 in the experiments.

\subsubsection{Parameter Estimation from Weighted Samples}
Based on the samples $l_j$ and weights $v_j$, the maximum likelihood estimation $\hat{\Lambda}$ is the solution to the following equation:
\begin{equation}
    \hat{\Lambda} = \frac{4}{\sum_j v_j} \sum_j \frac{v_j l_j l_j^\text{T}}{l_j^\text{T} \hat{\Lambda}^{-1} l_j}.
    \label{acgmle}
\end{equation}
The solution to Eq.~(\ref{acgmle}) can be computed by fixed-point iteration~\cite{ACG}. The final parameters of the updated proposal is determined the same way as in Eq.~(\ref{acgparam}).

\section{Details on Derivative Regularization Loss}

As stated in the main paper, the derivative regularization loss $\mathcal{L}_\text{reg}$ consists of the position loss $\mathcal{L}_\text{pos}$ and the orientation loss $\mathcal{L}_\text{orient}$. 

For $\mathcal{L}_\text{pos}$, we adopt the smooth L1 loss based on the Euclidean distance $d_t = \| t^\ast + \Delta t - t_\text{gt} \|$, given by:
\begin{equation}
    \mathcal{L}_\text{pos} = 
    \begin{dcases}
    \frac{d_t^2}{2\beta}, & d_t \leq \beta,\\
    d_t - 0.5\beta, & d_t > \beta,
    \end{dcases}
\end{equation}
with the hyperparameter $\beta$.

For $\mathcal{L}_\text{orient}$, we adopt the cosine similarity loss based on the angular distance $d_\theta$. For 1D orientation parameterized by the angle $\theta$, $d_\theta = \theta^\ast + \Delta \theta - \theta_\text{gt}$. For 3D orientation parameterized by the quaternion vector $l$, $d_\theta = 2 \arccos{(l^\ast + \Delta l)^\text{T} l_\text{gt}}$. The loss function is therefore defined as:
\begin{equation}
    \mathcal{L}_\text{orient} = 1 - \cos{d_\theta}.
    \label{orientloss}
\end{equation}
For 3D orientation, after the substitution, the loss function can be simplified to:
\begin{equation}
    \mathcal{L}_\text{orient} = 2 - 2 \left( (l^\ast + \Delta l)^\text{T} l_\text{gt} \right)^2.
\end{equation}

For the specific settings of the hyperparameter $\beta$ and loss weights, please refer to the experiment configuration code.

\section{Details on the Deformable Correspondence Network}

\subsection{Network Architecture} \label{deformnetsup}
\begin{figure}[t]
\begin{center}
    \includegraphics[width=0.95\linewidth]{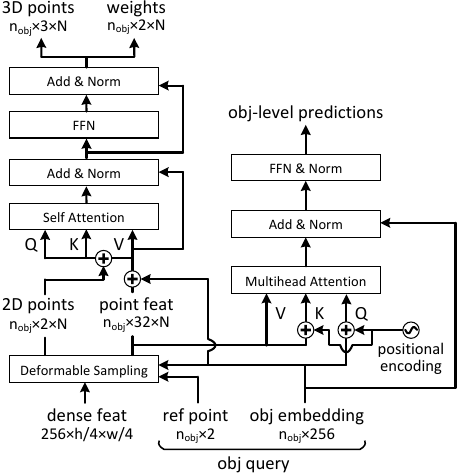}
\end{center}
\vspace{-1ex}
\caption{Detailed architecture of the deformable correspondence network.}
\label{fig:deformcorr}
\end{figure}

The detailed network architecture of the deformable correspondence network is shown in Figure~\ref{fig:deformcorr}.
% The design is largely inspired by the DETR family~\cite{detr,deformabledetr} of transformer-based object detectors. 
% In fact, the deformable correspondence network could have been integrated into the DETR framework. However, there was no DETR-like baseline for monocular 3D object detection before the very recent DETR3D~\cite{detr3d}, so we decide on the FCOS3D~\cite{fcos3d} framework as an alternative. 
Following deformable DETR~\cite{deformabledetr}, this paper adopts the multi-head deformable sampling. Let $n_\text{head}$ be the number of heads and $n_\text{hpts}$ be the number of points per head, a total number of $N = n_\text{head} n_\text{hpts}$ points are sampled for each object. The sampling locations relative to the reference point are predicted from the object embedding by a single layer of linear transformation. We set $n_\text{head}$ to 8, which yields $256 / n_\text{head} = 32$ channels for the point features.

The point-level branch on the left side of Figure~\ref{fig:deformcorr} is responsible for predicting the 3D points $x^\text{3D}_i$ and corresponding weights $w^\text{2D}_i$. The sampled point features are first enhanced by the object-level context, by adding the reshaped head-wise object embedding to the point features. Then, the features of the $N$ points are processed by the self-attention layer, for which the 2D points are transformed into positional encoding. The attention layer is followed by standard layers of normalization, skip connection, and feedforward network (FFN).

Regarding the object-level branch on the right side of Figure~\ref{fig:deformcorr}, a multi-head attention layer is employed to aggregate the sampled point features. Unlike the original deformable attention layer~\cite{deformabledetr} that predicts the attention weights by linear projection of the object embedding, we adopt the full Q-K dot-product attention with positional encoding. After being processed by the subsequent layers, the object-level features are finally transformed into to the object-level predictions, consisting of the 3D localization score, weight scale, 3D bounding box size, and other optional properties (velocity and attribute). Note that the attention layer is actually not a necessary component for object-level predictions, but rather a byproduct of the deformable point samples whose features can be leveraged with little computation overhead.

\subsection{Loss Functions for Object-Level Predictions}
As in FCOS3D~\cite{fcos3d}, we adopt smooth L1 regression loss for 3D box size and velocity, and cross-entropy classification loss for attribute. Additionally, a binary cross-entropy loss is imposed upon the 3D localization score, with the target $c_\text{tgt}$ defined as a score function of the position error:
\begin{align}
    c_\text{tgt} &= \mathit{Score}(\|t^\ast_\text{XZ} - {t_\text{XZ}}_\text{gt}\|) \notag\\
    &= \max(0, \min(1, -a \log{\|t^\ast_\text{XZ} - {t_\text{XZ}}_\text{gt}\|} + b)),
    \label{eqn:score}
\end{align}
where $t^\ast_\text{XZ}$ is the XZ components of the PnP solution, ${t_\text{XZ}}_\text{gt}$ is the XZ components of the true pose, and $a, b$ are the linear coefficients. The predicted 3D localization score $c_\text{pred}$ shall reflect the positional uncertainty of an object, as a faster alternative to evaluating the uncertainty via the Monte Carlo method during inference (Section~\ref{uncertainpose}). The final detection score is defined as the product of the predicted 3D score and the classification score from the base detector. 

\subsection{Auxiliary Loss Functions}

\subsubsection{Auxiliary Correspondence Loss}

\begin{figure}[t]
% \vspace{-2ex}
\begin{center}
    \includegraphics[width=1.0\linewidth]{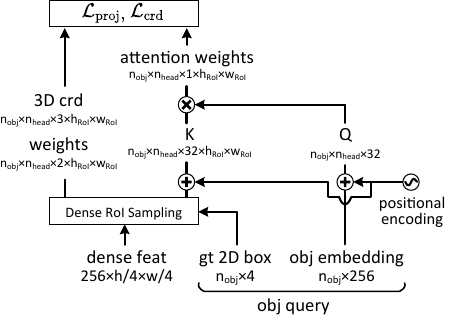}
\end{center}
\vspace{-1ex}
\caption{Architecture of the auxiliary branch. This branch shares the same weights of Q, K projection with the deformable attention layer in the lower right of Figure~\ref{fig:deformcorr}.}
\label{fig:auxiliary}
\end{figure}

To regularize the dense features, we append an auxiliary branch that predicts the multi-head dense 3D coordinates and corresponding weights, as shown in Figure~\ref{fig:auxiliary}. Leveraging the ground truth of object 2D boxes, the features within the box regions are densely sampled via RoI Align~\cite{maskrcnn}, and transformed into the 3D coordinates $x^\text{3D}$ and weights $w^\text{2D}$ via an independent linear layer. Besides, the attention weights $\phi$ are obtained via Q-K dot-product and normalized along the $n_\text{head}$ dimension and across the overlapping regions of multiple RoIs via Softmax. 

During training, we impose the reprojection-based auxiliary loss for the multi-head dense predictions, formulated as the negative log-likelihood (NLL) of the Gaussian mixture model~\cite{Bishop94mixturedensity}. Following \cite{monorun}, the reprojection error is further robustified by the Huber kernel $\rho(\cdot)$. The loss function for each sampled point is defined as:
\vspace{-1ex}
\begin{equation}
    \mathcal{L}_\text{proj} = \smash[b]{-\log \sum_{\text{RoI}}\sum_{k=1}^{n_\text{head}} \phi_k |\diag{w_k^\text{2D}}| \exp{-\frac{1}{2} \rho \left(\|f_k(y_\text{gt})\|^2 \right)}} ,
    \vspace{1mm}
    \label{auxreproj}
\end{equation}
where $k$ is the head index, $f_k(y_\text{gt})$ is the weighted reprojection error of the $k$-th head at the truth pose $y_\text{gt}$. In the above equation, the diagonal matrix $\diag{w_k^\text{2D}}$ is interpreted as the inverse square root of the covariance matrix of the normal distribution, i.e., $\diag{w_k^\text{2D}} = \Sigma^{-\frac{1}{2}}$, and the head attention weight $\phi_k$ is interpreted as the mixture component weight. $\sum_{\text{RoI}}$ is a special operation that takes the overlapping region of multiple RoIs into account, formulating a mixture of multiple heads and multiple RoIs (see code for details).

Another auxiliary loss is the coordinate regression loss that introduces the geometric knowledge. Following MonoRUn~\cite{monorun}, we extract the sparse ground truth of 3D coordinates $x^\text{3D}_\text{gt}$ from the 3D LiDAR point cloud. The robustified Gaussian mixture NLL loss for each sampled point with available ground truth is defined as:
\begin{equation}
    \mathcal{L}_\text{crd} = -\log{\sum_{k=1}^{n_\text{head}} \phi_k w^2 \exp -\frac{1}{2}  \rho\left(\left\| w \left(x^\text{3D}_k - x^\text{3D}_\text{gt}\right) \right\|^2\right)},
\end{equation}
where $w \in \mathbb{R}^+$ is a scalar weight parameter (bounded by $\exp$ activation) to be optimized during training. 

As with the KL loss in the main paper, dynamic loss weights~\cite{monorun} are employed to rescale these two auxiliary loss functions.

\subsubsection{Auxiliary Depth Loss}

For each projected LiDAR point in the image, we extract the feature vector from the interpolated dense feature map, which are then fed into a small 2-layer MLP to predict the scene depth. The output depth is represented by a Gaussian mixture distribution encoded by $\left\{{\phi_\text{D}}_k, z_k\right\}_{k=1}^{n_\text{D}}$, where the mixture weights ${\phi_\text{D}}_k$ are normalized via Softmax. Given the ground truth depth $z_\text{gt}$ of this point, the loss function is defined as:
\begin{equation}
    \mathcal{L}_\text{D} = -\log{\sum_{k=1}^{n_\text{D}} {\phi_\text{D}}_k w_\text{D} \exp -\frac{1}{2}  \rho\left( w_\text{D}^2 \left(z_k - z_\text{gt}\right)^2\right)},
\end{equation}
where $w_\text{D} \in \mathbb{R}^+$ is a weight parameter (bounded by $\exp$ activation) to be optimized during training.

\subsection{Training Strategy}

During training, we randomly sample 48 positive object queries from the FCOS3D~\cite{fcos3d} detector for each image, which limits the batch size of the deformable correspondence network to control the computation overhead of the Monte Carlo pose loss.

\subsection{Experiments on the Uncertainty of Object Pose}
\label{uncertainpose}
The entropy of the inferred pose distribution reflects the aleatoric uncertainty of the predicted pose. Previous work~\cite{monorun} reasons the pose uncertainty by propagating the reprojection uncertainty learned from a surrogate loss through the PnP operation, but that uncertainty requires calibration and is not reliable enough. In our work, the pose uncertainty is learned with the KL-divergence-based pose loss in an end-to-end manner, which is much more reliable fundamentally. 

To quantitatively evaluate the reliability of the pose uncertainty in terms of measuring the localization confidence, a straightforward approach is to compute the 3D localization score $c_\text{MC}$ via Monte Carlo pose sampling, and compare the resulting mAP against the standard implementation with 3D score $c_\text{pred}$ predicted from the object-level branch. With the PnP solution $t^\ast$, the sampled translation vector $t_j$, and its importance weight $v_j$, the Monte Carlo score is computed by:
\begin{equation}
    c_\text{MC} = \frac{1}{\sum_j v_j} \sum_j v_j \mathit{Score}\left(\|t^\ast_\text{XZ} - {t_\text{XZ}}_j\|\right),
\end{equation}
where the subscript $(\cdot)_\text{XZ}$ denotes taking the XZ components, and the function $\mathit{Score}(\cdot)$ is the same as in Eq.~\ref{eqn:score}.

As shown in Table~\ref{tab:addexp}, the mAP obtained via Monte Carlo scoring is on par with the standard implementation (0.393 vs. 0.392), indicating that the pose uncertainty is a reliable measure of the detection confidence. 

\begin{table}[ht]
\caption{Comparison between the scoring methods on the nuScenes validation set.}
\label{tab:addexp}
\centering
    \scalebox{1.0}{%
    \setlength{\tabcolsep}{0.5em}
    \begin{tabular}{lcccc}
        \toprule
        Scoring method & NDS\textuparrow & mAP\textuparrow & mATE\textdownarrow & mAOE\textdownarrow \\
        \midrule
        Standard & 0.463 & 0.392 & 0.626 & 0.282 \\
        Monte Carlo & 0.463 & 0.393 & 0.623 & 0.286 \\
        \bottomrule
    \end{tabular}}
\end{table}

\clearpage

\onecolumn
\section{Notation}

\begin{table}[h]
    \begin{center}
    \caption{A summary of frequently used notations.} 
    \label{tab:notation}
    \scalebox{1.0}{%
    \setlength{\tabcolsep}{0.5em}
    \begin{tabular}{cll}
        \toprule
        \multicolumn{2}{l}{Notation} & Description \\
        \midrule 
        $x^\text{3D}_i$ & $\in \mathbb{R}^3$ & Coordinate vector of the $i$-th 3D object point\\
        $x^\text{2D}_i$ & $\in \mathbb{R}^2$ & Coordinate vector of the $i$-th 2D image point\\
        $w^\text{2D}_i$ & $\in \mathbb{R}^2_+$ & Weight vector of the $i$-th 2D-3D point pair\\
        $X$ & & The set of weighted 2D-3D correspondences\\
        
        $y$ & & Object pose\\
        $y_\text{gt}$ & & Ground truth of object pose\\
        $y^\ast$ & & Object pose estimated by the PnP solver\\
        
        $R$ & & 3\texttimes3 rotation matrix representation of object orientation\\
        $\theta$ & & 1D yaw angle representation of object orientation\\
        % $\theta^\ast$ & & 1D yaw angle estimated by the PnP solver \\
        $l$ & & Unit quaternion representation of object orientation\\
        % $l^\ast$ & & Unit quaternion estimated by the PnP solver\\
        $t$ & $\in \mathbb{R}^3$ & Translation vector representation of object position\\
        $\Sigma_{y^\ast}$ & & Pose covariance estimated by the PnP solver \\
        
        $J$ & & Jacobian matrix\\
        $\tilde{J}$ & & Rescaled Jacobian matrix\\
        $F$ & & Concatenated vector of weighted reprojection errors of all points\\
        $\tilde{F}$ & & Concatenated vector of rescaled weighted reprojection errors of all points\\
        
        $\pi(\cdot)$ & $: \mathbb{R}^3 \rightarrow \mathbb{R}^2 $ & Camera  projection function\\
        $f_i(y)$ & $\in \mathbb{R}^2$ & Weighted reprojection error of the $i$-th correspondence at pose $y$\\
        $r_i(y)$ & $\in \mathbb{R}^2$ & Unweighted reprojection error of the $i$-th correspondence at pose $y$\\
        $\rho(\cdot)$ & & Huber kernel function\\
        $\rho^\prime_i$ & & The derivative of the Huber kernel function of the $i$-th correspondence\\
        $\delta$ & & The Huber threshold\\
        $p(X|y)$ & & Likelihood function of object pose\\
        $p(y)$ & & PDF of the prior pose distribution\\
        $p(y|X)$ & & PDF of the posterior pose distribution\\
        $t(y)$ & & PDF of the target pose distribution\\

        $q(y), q_t(y)$ & & PDF of the proposal pose distribution (of the $t$-th AMIS iteration)\\
        $y_j, y_j^t$ & & The $j$-th random pose sample (of the $t$-th AMIS iteration)\\
        $v_j, v_j^t$ & & Importance weight of the $j$-th pose sample (of the $t$-th AMIS iteration)\\
        
        $i$ & & Index of 2D-3D point pair\\
        $j$ & & Index of random pose sample\\
        $t$ & & Index of AMIS iteration\\
        $N$ & & Number of 2D-3D point pairs in total\\
        $K$ & & Number of pose samples in total\\
        $T$ & & Number of AMIS iterations\\
        $K^\prime$ & & Number of pose samples per AMIS iteration\\
        $n_\text{head}$ & & Number of heads in the deformable correspondence network\\
        $n_\text{hpts}$ & & Number of points per head in the deformable correspondence network\\
        $\mathcal{L}_\text{KL}$ & & KL divergence loss for object pose\\
        $\mathcal{L}_\text{tgt}$ & & The component of $\mathcal{L}_\text{KL}$ concerning the reprojection errors at target pose\\
        $\mathcal{L}_\text{pred}$ & & The component of $\mathcal{L}_\text{KL}$ concerning the reprojection errors over predicted pose\\
        $\mathcal{L}_\text{reg}$ & & Derivative regularization loss\\
        \bottomrule
    \end{tabular}}
    \end{center}
\end{table}

\clearpage
\twocolumn

\end{document}